\documentclass[graybox]{svmult}

\usepackage{type1cm}        
\usepackage{makeidx}         
\usepackage{graphicx}        
\usepackage{multicol}        
\usepackage[bottom]{footmisc}
\usepackage{booktabs}

\usepackage{newtxtext}       %
\usepackage[varvw]{newtxmath}       

\usepackage{subfig}
\usepackage{enumitem}
\usepackage{tabularx}
\usepackage{csquotes}
\usepackage{url}

\usepackage{xspace}
\newcommand{\totalsize}{101 million\xspace}
\newcommand{\filtersize}{11.7 million\xspace}
\newcommand{\archsize}{2.6 million\xspace}

\usepackage{xcolor}
\usepackage{etoolbox}
\definecolor{cb-green-sea}  {RGB}{  0, 146, 146}
\definecolor{cb-blue}       {RGB}{ 0, 109, 219}
\definecolor{cb-burgundy}   {RGB}{146,   0,   0}

\newcommand{\Dbbb}{\ensuremath{\bullet\!\!\bullet\!\!\bullet}}
\newcommand{\Dbb}{\ensuremath{\bullet\!\bullet}}
\newcommand{\Db}{\ensuremath{\bullet}}
\newcommand{\Dnb}{\ensuremath{-}}

\makeindex             

\begin{document}

\title*{Generative AI and the History of Architecture}
\author{Joern Ploennigs\orcidID{0000-0002-6320-8891} and\\ Markus Berger\orcidID{0000-0001-5232-0039}}
\institute{Joern Ploennigs \and Markus Berger \at AI for Sustainable Construction, University of Rostock, Germany\newline
\email{{Joern.Ploennigs, Markus.Berger}@uni-rostock.de}}
%

\maketitle

\abstract{
Recent generative AI platforms are able to create texts or impressive images from simple text prompts. This makes them powerful tools for summarizing knowledge about architectural history or deriving new creative work in early design tasks like ideation, sketching and modelling. But, how good is the understanding of the generative AI models of the history of architecture? Has it learned to properly distinguish styles, or is it hallucinating information? In this chapter, we investigate this question for generative AI platforms for text and image generation for different architectural styles, to understand the capabilities and boundaries of knowledge of those tools. We also analyze how they are already being used by analyzing a data set of \totalsize Midjourney queries to see if and how practitioners are already querying for specific architectural concepts.}

\section{Introduction}\label{sec:intro}

Throughout the history of architectural design, various styles developed and were memorialized in our built environment. Many of them are still used as an inspiration for new designs or are fused into new styles within modern architecture. Entangled with the styles are technical innovations in how we design buildings, from traditional manual sketches to modern CAD drawings and 3D renderings. Most recently a new kind of generative AI tools is creating new ways to design and illustrate at scale using those historical styles. Generative AI models for text like ChatGPT allow us to summarize the comprehensive knowledge about historical styles available on the internet and instantly derive a multitude of different representations, from short descriptions to lists of attributes. Image generation models like Midjourney on the other hand enable a new productivity in visually exploring, re-shaping and fusing historical styles. Both model types are trained on massive libraries of texts and images, which include examples of every prominent historical style, leading to a re-appearance and renewal of these styles in their output.

Within the chapter we will analyze how these generative AI tools can and are used for analysing and incorporating historical architectural styles into new designs. We start by reviewing generative AI technologies from text generation models to image generation models. We then investigate both technologies in their knowledge of historical architectural designs and how prone they are to hallucination. Finally we investigate the practical application of those technologies by analysing more than \totalsize queries from the Midjourney platform to determine frequency and usage patterns in generating architectural images. Particularly, we will quantitatively explore which historical architectural styles and epochs commonly appear in queries, which architects are invoked together with these styles, and how styles are combined in an attempt to create unique designs. In a further quantitative analysis we will also explore which architectural styles the model can recognize when reversing the generation process from image to prompt.

The chapter will have multiple novel scientific contributions including:
\begin{itemize}
\item A discussion of the technology of ChatGPT and Midjourney as exemplary generative AI platforms
\item An analysis of the knowledge of historical styles in ChatGPT
\item An analysis of how prone ChatGPT is to hallucinating facts for styles
\item An analysis of the knowledge of historical styles in Midjourney
\item An analysis of how well Midjourney encodes the characteristics of styles
\item An analysis of which architectural styles and architects are popular in Midjourney
\end{itemize}

\section{State of the Art in Generative Methods}\label{sec:soa}

The recent surge in research and development into generative AI has shown a large amount of potential application domains. There are successful large-scale platforms that encapsulate the process of model training and usage, like ChatGPT and Midjourney. There are research papers using both the models themselves, as well as the platforms to solve specific problems. Text-to-Text models like ChatGPT are used for project management \cite{prieto2023investigating}, worker safety \cite{uddin2023leveraging}, or portfolio optimization \cite{Donncha23}. Text-to-image models are used for generating images for concept design \cite{cheng2023concept}, urban planning \cite{seneviratne2022dalle}, floor plan design \cite{ploennigs2023diffusion}, historic reconstruction \cite{moral2023can}, or education \cite{Berger23}.

These recent generative successes are driven by decades of research into sub-disciplines of the field of deep learning, like image understanding and computer vision on the imaging side, and \textit{Natural Language Processing (NLP)} on the textual side. The first big step towards generative methods was \cite{Vaswani17}, that introduced the transformer architecture that we still use in recent models. It revolutionized the capability of deep learning NLP models to understand and utilize language, by training models to work on a larger context than simple word neighborhood. Like generative adversarial networks (GANs), they consist of an encoder and a decoder \cite{goodfellow2014generative} working in tandem to enable the transformation of text into a highly effective internal representation, and then back into text. The benefit of transformer approaches is that it is easy to parallelize the computation on GPUs allowing them to train massive amounts of training data that allows for large internal parameter counts \cite{brown2020language, ouyang2022training}.

This original invention allowed the development of \textit{Large Language Models (LLM)} like GPT \cite{radford2018improving}, BERT \cite{devlin2018bert}, BART \cite{lewis2019bart}, and T5 \cite{raffel2020exploring}, all based on the transformer architecture. Quickly, the capability of transformers was extended beyond just text, with applications in the image generation domain like VisualBERT \cite{li2019visualbert}. Those \textit{Vision Language Models (VLM)} are trained on labeled images and predict the next cluster of pixels by utilizing another encoder that turns an image into a sequence of word tokens and vice versa.

Another technique that developed starting with \cite{sohl2015deep}, were so-called diffusion methods, in which models are trained by slowly destroying information in high-quality images and training them to reverse this process, allowing them to take a low-quality image (or even random noise) and subsequently generate new information that turns it into an image of much higher quality.

By combining this diffusion method with large language models, this process of reverse diffusion could be driven by user-defined textual input prompts \cite{radford2021learning, ramesh2021zero, Ramesh22}. This step moved the technology from lab research into a technology that users could immediately experiment with.

With the technology being this attainable, several platforms rose to fame even in the general public. This includes DALL-E, ChatGPT, Midjourney, Bard and StableDiffusion, each offering unique combinations of text and image generation methods. Like all deep learning, the capabilities and characteristicts of these models are distinctly defined by their training datasets, the extent and contents of which are often unknown. One known example is StableDiffusion, which was trained on the LAION-5B \cite{Rombach_2022_CVPR} training dataset.

New approaches or combinations of existing ideas are constantly enabling new kinds of AI-driven workflows. Recently, there has been progress in generating video \cite{ho2022video}, 3D models via point clouds \cite{luo2021diffusion,zeng2022lion}, and even 3D animation data \cite{tevet2022human}, usually based on similar architectures to the image models or even directly involving an image model. While this is beyond the scope of this chapter, especially the generation of 3D models is a big opportunity for architecture, as early work is showing \cite{sebestyengenerating}. 




While those generative AI models are rapidly improving and opening the potential for new applications, the fundamental workflows and challenges like hallucination are here to stay. For this reason, we will to investigate two of the largest currently available platforms for text and image generation for one specific application domain: \textit{the history of architecture}.

\section{Analysis of Architectural Knowledge of Generative AIs}\label{sec:knowhow}
\subsection{Knowledge in Text Generation with ChatGPT}
\subsubsection{Summarizing Architectural History with ChatGPT around the Globe}

\begin{figure}[ht]%
\centering
\includegraphics[width=\textwidth]{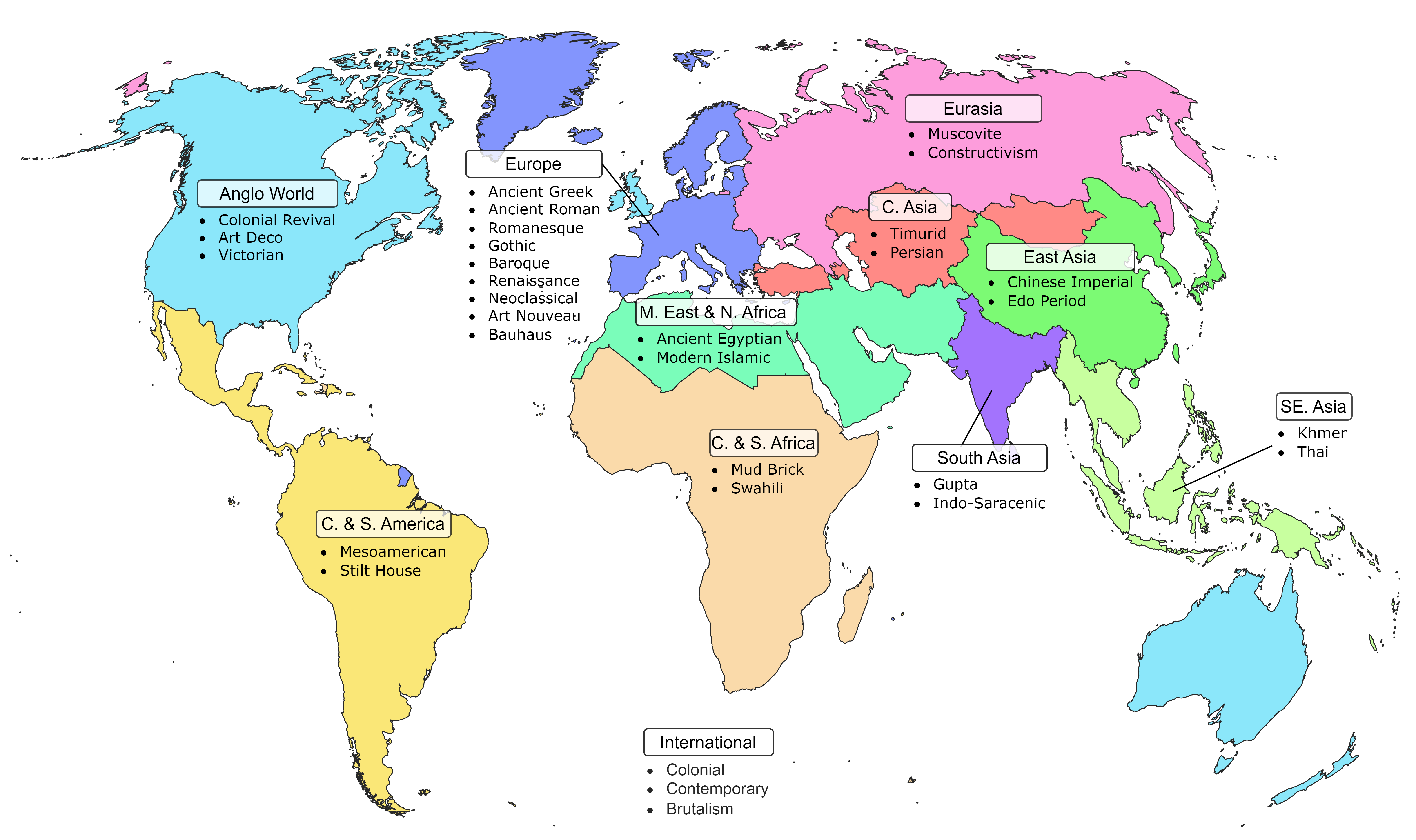}
\caption{Our selection of socio-cultural regions of the world, based on \cite{regions2019} together with a list of the most relevant architectural styles as responded by ChatGPT v3.5.}
\label{fig:world_regions}
\end{figure}

ChatGPT is a generative AI that allows generate texts in response to queries. To check its general knowledge about architectural styles, we asked it for a list of the most important historical architectural styles  10 times. We ranked that list and retrieved a list of well known styles from Ancient Greek, over Gothic to modern styles like Contemporary. Those styles originated primarily from western history. This is not surprising as they are well documented in the English speaking internet on which ChatGPT was trained. To increase the challenge and verify if it actually understands geographical and cultural context, we repeated that process for different cultural regions. This selection of regions is based on the country similarity index from \cite{regions2019}, that we further adjusted for architecture, by merging regions that historically used similar styles. Fig.~\ref{fig:world_regions} shows a map of these cultural regions with their respective styles. For every region, we asked ChatGPT v3.5 for the top 3 architectural styles 10 times. With this approach we collected 111 styles from various regions and periods. We down-selected those to the most frequent 30 styles listed in Fig.~\ref{fig:world_regions} to retrieve a sample set of well known historic styles as well as less known styles like Mud Brick or Muscovite for our further analysis.

We then asked ChatGPT for each style to provide a detailed description, time period, characteristics, famous architects, and exemplar constructions as JSON. For each architect and example we asked for further details like time range, location, and description. We also asked for a summary of the styles in nouns and adjectives that we will later use for generating images in Midjourney. ChatGPT provided data for all styles, without ever stating that it does not know enough about a particular style.

\begin{figure}[t]
    \centering
    \subfloat[\centering ChatGPT v3.5 \label{fig:describe_flow_chatgpt}]{{\includegraphics[width=0.49\textwidth]{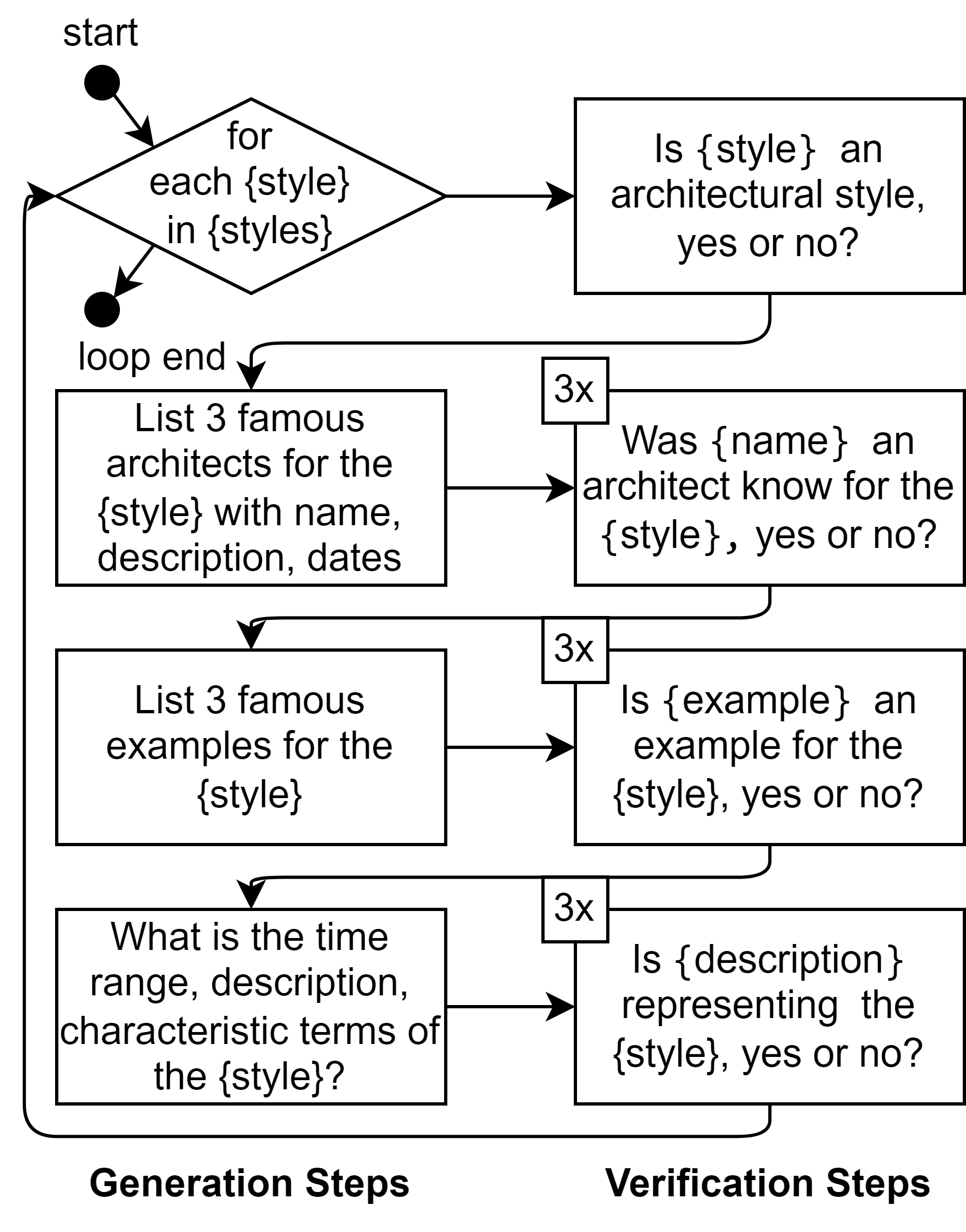} }}%
    \subfloat[\centering Midjourney \label{fig:describe_flow_midjourney}]{{\includegraphics[width=0.49\textwidth]{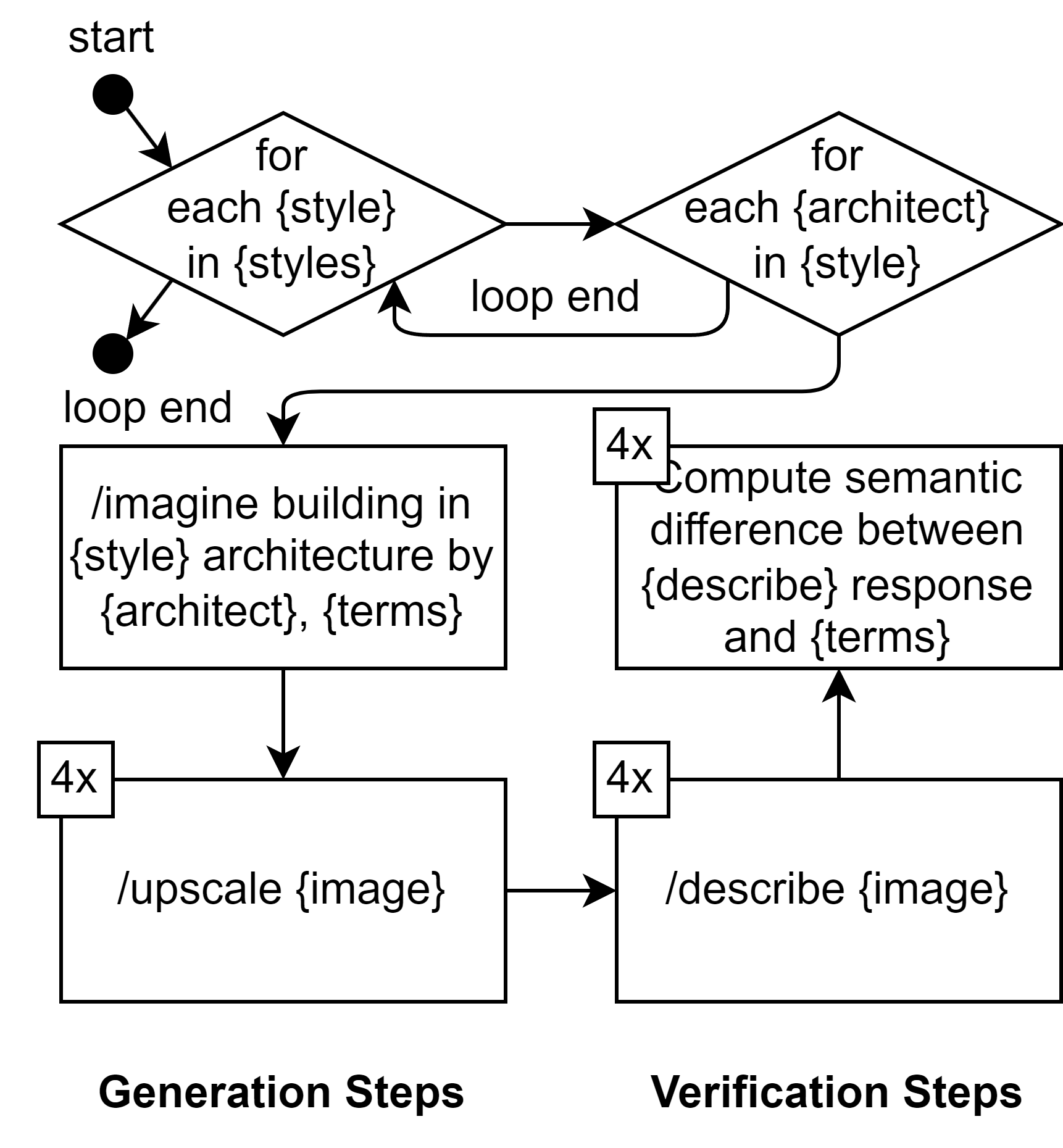} }}%
    \caption{Process to generate and verify AI content}
    \label{fig:describe_flow}%
\end{figure}

\subsubsection{Is Hallucination a Problem?}

This hints at a common challenge with generative AIs like ChatGPT: they will make up missing information, commonly known as \textit{hallucination} \cite{bang2023multitask}. To verify the correctness of the information we took two approaches. We first checked with control questions, whether ChatGPT does confirm its own answer. For example we asked with deactivated query memory: \enquote{Is \{style\} an architectural style? Answer with yes or no.} or \enquote{Was \{name\} an architect know for the \{style\} architectural style? Answer with yes or no.}. We repeated that 3 times and computed a \textit{confidence score} of how often ChatGPT answered Yes out of those. Secondly, we manually verified whether the generated information is correct and evaluated each of the selected 30 styles with a \textit{quality score} between 0 to 1, based on information available in literature or the internet. This allowed us to give 1/3 or 2/3 points if ChatGPT responded partially correct results, like naming a king, who was not specifically an architect, but, still from the period. Fig.~\ref{fig:describe_flow_chatgpt} shows our whole process of generating and verifying the knowledge with ChatGPT.

Table~\ref{tab:styles} lists the results of this analysis for all architectural styles. For better readability we represent the numbers for the confidence and quality scores with \Dbbb\xspace if 100\,\%, \Dbb\xspace if between 99--66\,\%, \Db\xspace if between 65--33\,\%; and \Dnb\xspace if between 32--0\,\%. The first question we evaluated is: \enquote{Is \{style\} an architectural style?} (\textit{Existence}). It is notable, that the confidence of ChatGPT itself is actually very low in most cases, but we could still validate most of the styles. We lowered the rank some styles like \enquote{Swahili}, \enquote{Mud Brick}, \enquote{Stilt House}, and \enquote{Timurid Period}, as it is debatable whether they are an architectural style or rather a construction method. However, they are all culturally representative for their respective regions, making them similar to styles for purposes of image generation and understanding. The quality of the style \textit{descriptions} that we manually evaluated was high in almost all cases and succeeds in providing characteristic information for each style. Also ChatGPTs own confidence of those descriptions is high, which we validated with the control question about whether the description represents the requested style. The confidence of the representative architects for each styles is actually matching our own quality evaluation. Meaning, ChatGPT invented several architects that it then did not confirm in the control question particularly for styles for which we don't know architect names like \enquote{Gupta} or \enquote{Khmer} or which do not translate well to English like architects from the \enquote{Edo} or \enquote{Chinese Imperial} periods. Interestingly this does not apply to \enquote{Egyptian} or \enquote{Greek} architects, as architects are well documented in their English translations. ChatGPT was way more reliable in listing relevant examples for each style and only failed for styles like \enquote{Swahili} or \enquote{Stilt House}, where examples are not clearly assigned.

This evaluation shows that ChatGPT has a broad knowledge of architectural styles including African or Asian styles that are less well documented in many English speaking sources. On the other hand, it shows that it is not that reliable on naming architects or examples for less well documented styles and will hallucinate and invent names including biographies of those architects. It also shows that using control questions to compute a confidence score allowed us to use ChatGPT to quickly verify the correctness of its own information.

\begin{table}[]
\scriptsize
\centering
\begin{tabular}{@{}llccccccccccc@{}}
\toprule
Cultural & Architectural & \multicolumn{2}{c}{Period} & \multicolumn{2}{c}{Existence} & \multicolumn{2}{c}{Description} & \multicolumn{2}{c}{Architects} & \multicolumn{2}{c}{Examples} \\
Area & Style & Start & End & Conf. & Quality & Conf. & Quality & Conf. & Quality & Conf. & Quality \\ \midrule
C. \& S. America & Mesoamerican & 2000 BC & 1519 CE & \Dnb & \Dbb & \Dbb & \Dbbb & \Db & \Db & \Dbbb & \Dbbb \\
          & Stilt House & 6000 BC & present & \Dnb & \Db & \Dbbb & \Dbbb & \Dnb & \Dnb & \Dnb & \Dnb \\
Anglo World & Colonial Revival & 1880 CE & 1930 CE & \Dbbb & \Dbb & \Dbbb & \Dbbb & \Dbb & \Dbbb & \Dbb & \Db \\
          & Victorian & 1837 CE & 1901 CE & \Dbb & \Dbb & \Dbbb & \Dbbb & \Dbbb & \Dbbb & \Dbb & \Dbbb \\
          & Art Deco & 1920 CE & 1939 CE & \Dbbb & \Dbbb & \Dbbb & \Dbbb & \Dbb & \Dbbb & \Dbbb & \Dbbb \\
Intercultural & Colonial & 1600 CE & 1947 CE & \Dnb & \Dbb & \Dbbb & \Dbb & \Dnb & \Db & \Dbbb & \Db \\
          & Brutalism & 1950 CE & 1970 CE & \Dbbb & \Dbbb & \Dbbb & \Dbbb & \Dbb & \Dbbb & \Dbbb & \Dbb \\
          & Contemporary & 1950 CE & present & \Dbbb & \Dbbb & \Dbbb & \Dbbb & \Dbbb & \Dbbb & \Dbb & \Dbb \\
Europe & Ancient Greek & 900 BC & 146 BC & \Dnb & \Dbbb & \Dbbb & \Dbbb & \Dbb & \Dbbb & \Dbbb & \Dbbb \\
          & Ancient Roman & 300 BC & 476 CE & \Db & \Dbbb & \Dbbb & \Dbbb & \Dbbb & \Dbbb & \Dbbb & \Dbbb \\
          & Romanesque & 800 CE & 1200 CE & \Dbbb & \Dbbb & \Dbbb & \Dbbb & \Dnb & \Dbbb & \Dbbb & \Dbb \\
          & Gothic & 1100 CE & 1500 CE & \Dbbb & \Dbbb & \Dbbb & \Dbbb & \Dbb & \Dbb & \Dbbb & \Dbbb \\
          & Baroque & 1584 CE & 1750 CE & \Dbbb & \Dbbb & \Dbbb & \Dbbb & \Dbbb & \Dbbb & \Dbb & \Dbb \\
          & Renaissance & 1400 CE & 1600 CE & \Dnb & \Dbbb & \Dbbb & \Dbbb & \Dbbb & \Dbbb & \Dbbb & \Dbbb \\
          & Neoclassical & 1740 CE & 1909 CE & \Dbbb & \Dbbb & \Dbbb & \Dbbb & \Dbb & \Dbb & \Dbbb & \Dbbb \\
          & Art Nouveau & 1890 CE & 1914 CE & \Dbbb & \Dbbb & \Dbbb & \Dbbb & \Dbbb & \Dbbb & \Dbbb & \Dbbb \\
          & Bauhaus & 1919 CE & 1933 CE & \Dbbb & \Dbbb & \Dbbb & \Dbbb & \Dbb & \Dbbb & \Dbb & \Dbbb \\
Eurasia & Muscovite & 1407 CE & 1692 CE & \Dnb & \Dbbb & \Dbb & \Dbbb & \Db & \Dbb & \Dbbb & \Dbbb \\
          & Constructivism & 1917 CE & 1930 CE & \Dbbb & \Dbbb & \Dbb & \Dbbb & \Dbbb & \Dbbb & \Dbbb & \Dbbb \\
C. \& S. Africa & Mud Brick & 4000 BC & present & \Dnb & \Db & \Dbb & \Dbbb & \Dbb & \Db & \Dbb & \Dbbb \\
          & Swahili & 800 CE & 1800 CE & \Dnb & \Db & \Dbbb & \Dbbb & \Dnb & \Dnb & \Dbbb & \Db \\
M. East \& N. Africa & Ancient Egyptian & 3100 BC & 30 BC & \Dbb & \Dbbb & \Dbbb & \Dbbb & \Dbb & \Dbbb & \Dbbb & \Dbbb \\
          & Modern Islamic & 1922 CE & present & \Dnb & \Dbb & \Dbbb & \Dbbb & \Dbb & \Dnb & \Dbb & \Dbbb \\
S. Asia & Gupta & 320 CE & 550 CE & \Dnb & \Dbb & \Dbbb & \Dbbb & \Dnb & \Db & \Dnb & \Db \\
          & Indo-Saracenic & 1795 CE & 1947 CE & \Dbbb & \Dbbb & \Dbbb & \Dbbb & \Dbb & \Dbbb & \Dbbb & \Dbbb \\
E. Asia & Chinese Imperial & 618 CE & 1912 CE & \Dnb & \Dbbb & \Dbbb & \Dbbb & \Dnb & \Dnb & \Dbbb & \Db \\
          & Edo Period & 1603 CE & 1657 CE & \Dnb & \Dbb & \Dnb & \Dbbb & \Dnb & \Dnb & \Dbb & \Db \\
SE. Asia & Khmer Empire & 825 CE & 1431 CE & \Dnb & \Dbb & \Dbbb & \Dbbb & \Dnb & \Db & \Dbb & \Dbbb \\
          & Thai & 1000 CE & present & \Dnb & \Dbb & \Dbbb & \Dbbb & \Dnb & \Dnb & \Dbbb & \Dbbb \\
C. Asia & Timurid Period & 1370 CE & 1507 CE & \Dnb & \Db & \Dbbb & \Dbbb & \Dnb & \Dnb & \Dbbb & \Dbbb \\
          & Persian & 600 BC & 600 CE & \Dnb & \Dbbb & \Dbbb & \Dbb & \Db & \Dbb & \Dbb & \Dbb \\
\bottomrule
\end{tabular}
    \caption{Evaluation of Architectural Styles in ChatGPT v3.5 with \Dbbb: 100\,\%, \Dbb: 99--66\,\%; \Db: 65--33\,\%; \Dnb: 32--0\,\% }%
    \label{tab:styles}%
\end{table}

\subsubsection{Semantic Similarity of Architectural Styles}

\begin{figure}[t]%
    \centering
    \subfloat[\centering Style and Characteristic Words]{{\includegraphics[width=0.49\textwidth]{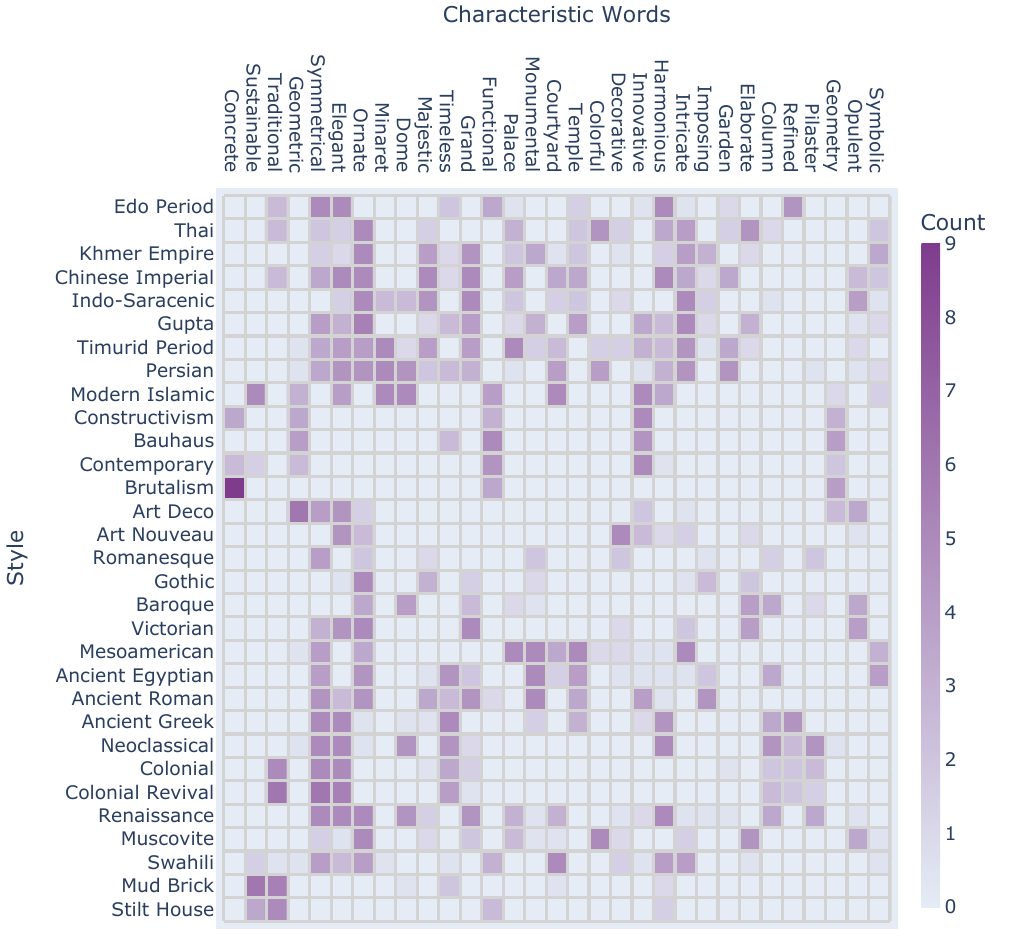} }}%
    \subfloat[\centering Style Similarity]{{\includegraphics[width=0.49\textwidth]{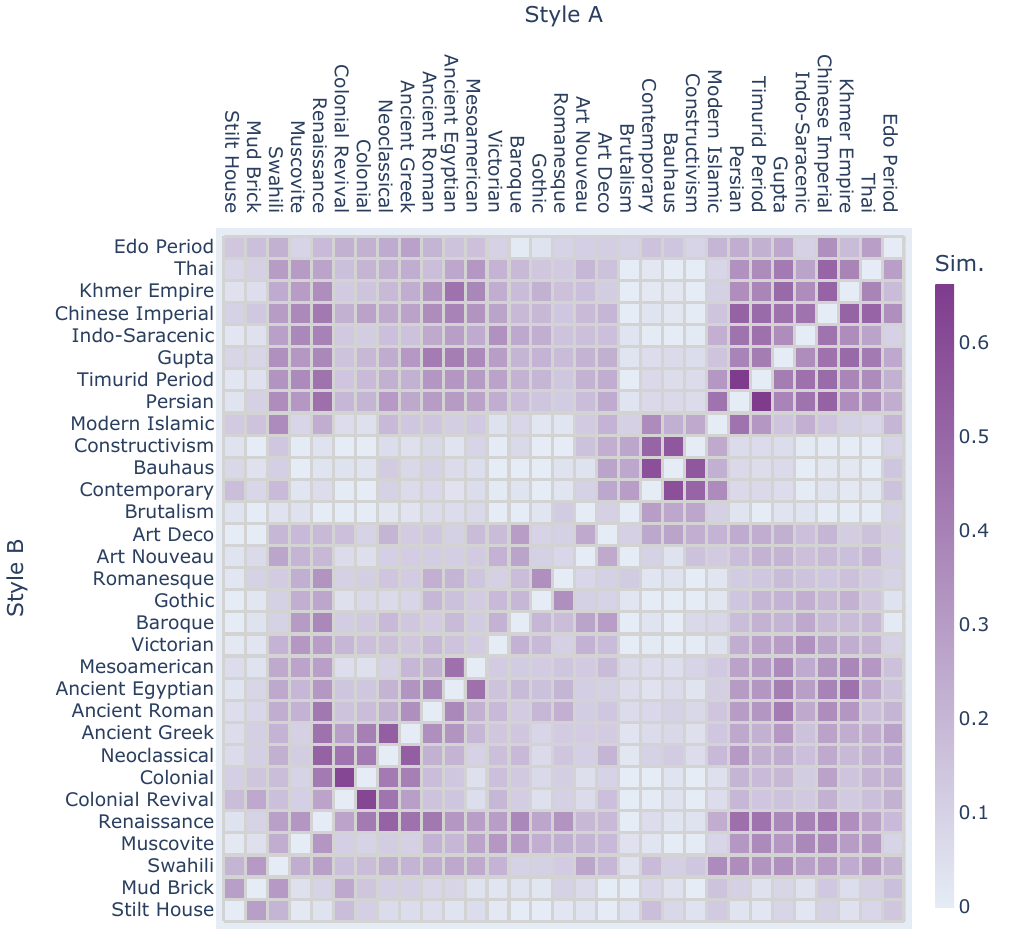} }}%
    \caption{Semantic Similarity between Styles}%
    \label{fig:archi_styles}%
\end{figure}

In the next step we were interested in comparing the semantic similarity of the styles based on the characteristic descriptions created by ChatGPT. Similar analyses were done in the past using the information in Wikipedia and DBpedia \cite{ruiz2019potential,mille2020case} focusing strongly on western styles. As shown before, ChatGPT has a solid knowledge of styles from different cultural regions, allowing us to analyse similarities for a larger variety of styles. We compute a semantic similarity score. First, we asked ChatGPT 10 times to describe each style with characteristic nouns and adjectives and computed a frequency of those words, to get a ranking of the most important ones and reduce the influence of hallucination. Fig.~\ref{fig:archi_styles} a) shows the frequency of the most common words along with their styles. For example, \enquote{Concrete} is very characteristic for \enquote{Brutalism} and also used for \enquote{Contemporary} and \enquote{Constructivism}, but no other style. \enquote{Functional} is used for all modern styles, but, never \enquote{Ornate},\enquote{Symmetrical}, or \enquote{Elegant}. \enquote{Refined}, \enquote{Plaster}, and \enquote{Column} are used for characterising \enquote{Ancient Greek}, \enquote{Neoclassical}, \enquote{Colonial (Revial)}, and \enquote{Renaissance}.

Next, we computed a \textit{semantic similarity score} between two styles to cluster them. Therefore, we intersected the characteristic words of two styles $a$ and $b$ and weighted them by their combined frequency normalized by the total frequency. Meaning if a word is frequently used to characterise style $a$ and less frequently used to characterise style $b$ it gets weighted less than a word that is frequently used for both. Fig.~\ref{fig:archi_styles} b) shows the similarity matrix after spectral co-clustering \cite{dhillon2001co}. The similarity ranges from 0--60\,\% for styles that share many words with high frequency in their characterization by ChatGPT. There are multiple clusters identifiable in the figure. First, we have the cluster of Asian styles in the top right corner. The \enquote{Chinese Imperial} style shares the highest semantic similarity with most of those styles. Interestingly, this also included Arabic styles like \enquote{Persian} and the \enquote{Timurid Period} that form the next small sub-cluster. The next larger isolated cluster is formed by the modern styles of \enquote{Contemporary}, \enquote{Bauhaus}, \enquote{Constructivism}, and \enquote{Brutalism}. It is notable, that this is also the least connected cluster. This means that the next big cluster distantly related to \enquote{Renaissance} shares more semantic similarity with Asian styles than with Modern ones. This \enquote{Renaissance} cluster splits into several smaller clusters. First there are several pairs of styles from connected time periods like \enquote{Art Deco} and \enquote{Art Nouveau} or \enquote{Romanesque} and \enquote{Gothic}. It is interesting that the latter two do not share strong similarities with other styles. The only style that is sharing semantic similarity is the \enquote{Renaissance}, which is actually forming the next big cluster. It stretches from \enquote{Ancient Greek} to \enquote{Neoclassic} and \enquote{Colonial} and \enquote{Colonial Revival}. Common in all those styles are simple columns. The last cluster is formed by rural styles like \enquote{Mud Brick} and \enquote{Stilt House}.

\subsection{Knowledge in Image Generation with Midjourney}

\subsubsection{How Well Does Midjourney Recognize its Own Style Representations?}

Midjourney is a generative AI that is capable of generating images from text prompts (text-to-image). One of its newer capabilities is the ability to describe uploaded images with text. Specifically, this means that the model is capable of image-to-text transformation. We utilize both features to take a closer look at how well Midjourney can recognize and distinguish specific architectural styles. Particularly, how well it can recognize the style in its own images, by feeding model outputs back into the describe feature. This allows us to check if it actually knows the characteristics of styles or is hallucinating them, similar to our analysis of ChatGPT.

Therefore, we used the same list of architectural styles for different cultural zones that we generated with ChatGPT, including the architects, and generated characteristic terms for each style. From the style names and the descriptive terms we computed a set of prompts, based on the template \enquote{Building in \{name\} architecture by \{architect\}, \{terms\}}. With these prompts we give Midjourney multiple avenues to understand what should be visible in the image from the style name, to the architects, and description terms. 

We entered each prompt into Midjourney v5 three times with a different architect, yielding 12 images for each of the 30 styles. Each of these images was upscaled once with the basic upscaling model. The resulting 360 images were then used as inputs for the img2txt-model, also called the \enquote{describe}-feature. This feature generates four prompts that try to describe the input image in different ways. This means that at the end of the process we are left with 48 descriptions for each style (all based on the same prompt), for a total of 1440 data points. Fig.~\ref{fig:describe_flow_midjourney} illustrates this process.

\begin{figure}[t]%
    \centering
    \subfloat[\centering Occurrences of names and terms\label{fig:midjourney_describe_occur}]{{\includegraphics[width=0.45\textwidth]{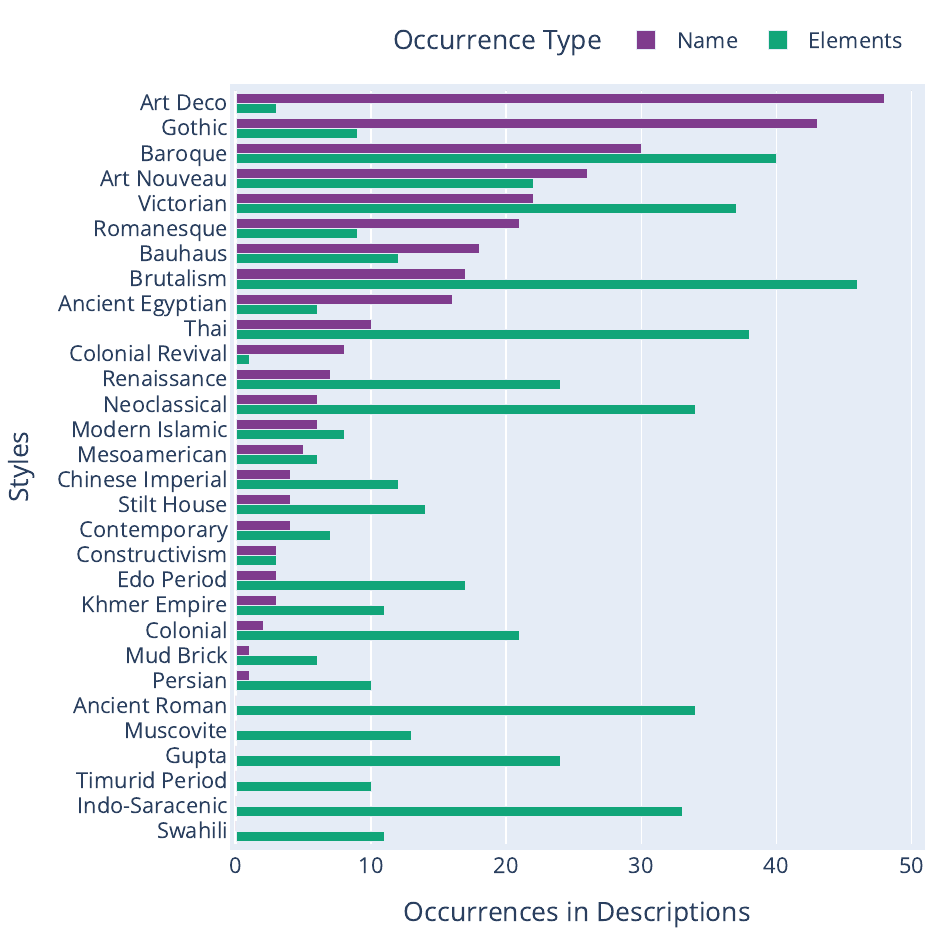} }}%
    \subfloat[\centering Misattributions of styles\label{fig:midjourney_describe_misattr}]{{\includegraphics[width=0.55\textwidth]{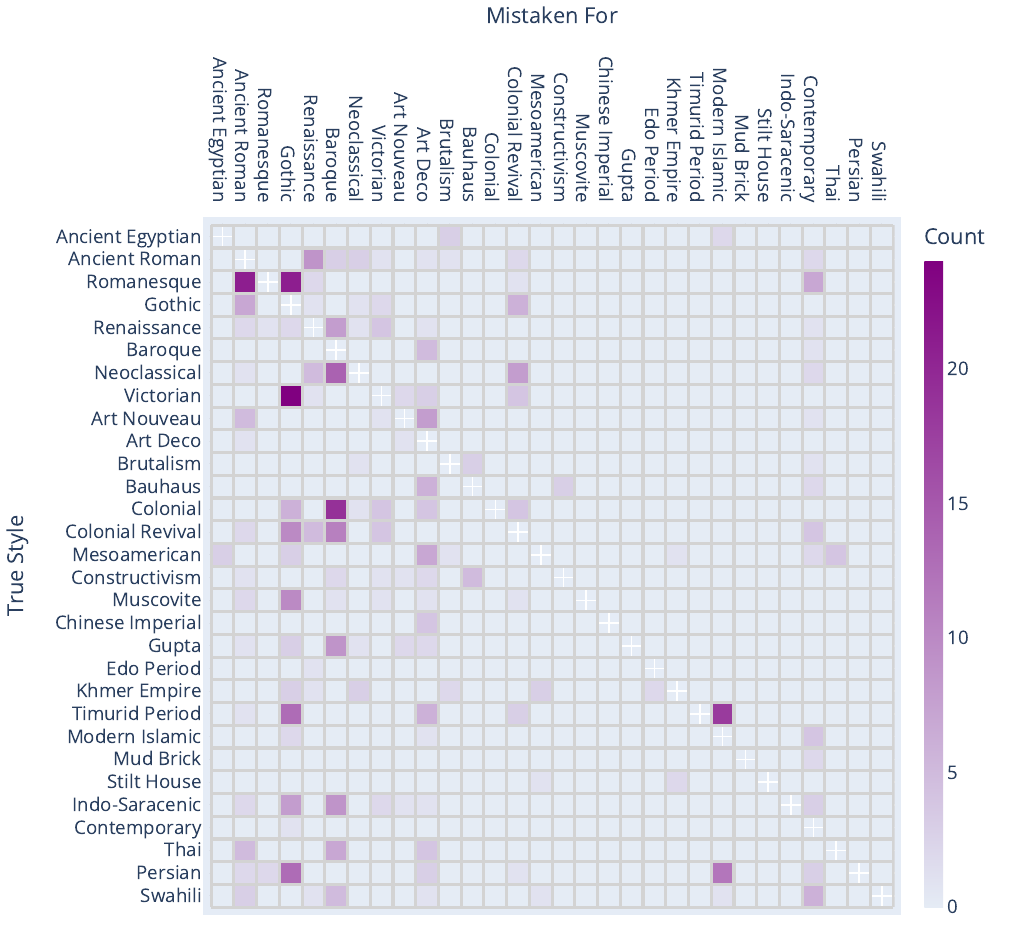} }}%
    \caption{Identifyability of architectural styles with the describe-feature in Midjourney}%
    \label{fig:midjourney_describe}%
\end{figure}

We used this data for two analyses: First, we simply checked each description for whether the name of the style was correctly invoked. The name in this case is the relevant part of the name, for example \enquote{Khmer} instead of the full \enquote{Khmer Empire}. Each description that had a correct mention of the style was marked as an occurrence. Fig.~\ref{fig:midjourney_describe_occur} shows this count by style in purple. It is notable that the European and Anglo styles are easiest to identify for Midjourney, followed by certain Asian and African styles. Perhaps surprisingly, Midjourney was not able to identify the very iconic Roman style of architecture. The likely reason for this is that Roman influences are integrated so deeply into later European styles that the actual historical buildings are difficult to recognize as \enquote{the original}.

In a second step, we checked whether any of the important elements of the style that ChatGPT distinguished were used in the description. Every element was described by one or two words, and it was enough for one element to appear in the description. The green bars in Fig.~\ref{fig:midjourney_describe_occur} show that there is large variance in the styles, and no strong correlation with the occurrences of the style name. Some styles, it seems, are not strongly encoded by name, but are still present within the model by the common descriptors of their elements. This likely also comes back to the effectiveness of ChatGPT in describing these styles in the first place. For example, the colonial revival style has a low rating in both types of occurrences, likely because the \enquote{revival} was the relevant part of the style name, and it is difficult for all involved models to continually distinguish it from the original colonial style through every step of our generation process.

The second analysis in turn highlights not the successes, but the errors committed while describing the style generations. Fig.~\ref{fig:midjourney_describe_misattr} displays for each style, how often every other style appeared in the descriptions generated for it, i.e., how often each style was mistaken for each of the other styles. Here we see the likely reason for the poor performance in identifying the Ancient Roman style earlier: it is very often misattributed as the medieval, roman-inspired \enquote{Romanesque} style. 

Which architect was invoked for each generation sometimes made a difference in which kind of building or structure was shown, but there was almost no mention of architects in any of the descriptions. For this reason, we excluded the architects from the analysis, but kept them in the prompts for more varied images.

Other notable results include the high likelihood of the \enquote{Gothic} style to be confused with \enquote{Victorian}, which again is a revival style of the former. In fact, the vast influence the Gothic style had on contemporary and future styles can be seen by just how often other styles are confused for it, and how few misattributions it itself has. A less euro-centric data point of note is that the Timurid style is often confused for the Modern Islamic style, but not vice versa. This is most likely an effect of the Timurid style and construction method influencing Islamic architecture to this day, while not being very present within the model itself (as can be seen in Fig.~\ref{fig:midjourney_describe_occur}).

Apart from this quantitative analysis, there are also clear qualitative distinctions in how well Midjourney handles individual styles, which we explore next.

\subsubsection{Example Descriptions}

Art Deco is a relatively recent style that originated in Europe and then flourished in the Anglo World. ChatGPT used the following terms to describe it: \enquote{geometric shapes bold colors luxurious materials streamlined forms symmetrical patterns decorative motifs plush interiors lavish embellishments glamorous aesthetics modernist design}.

\begin{figure}[h]
    \centering
    \includegraphics[width=0.5\textwidth]{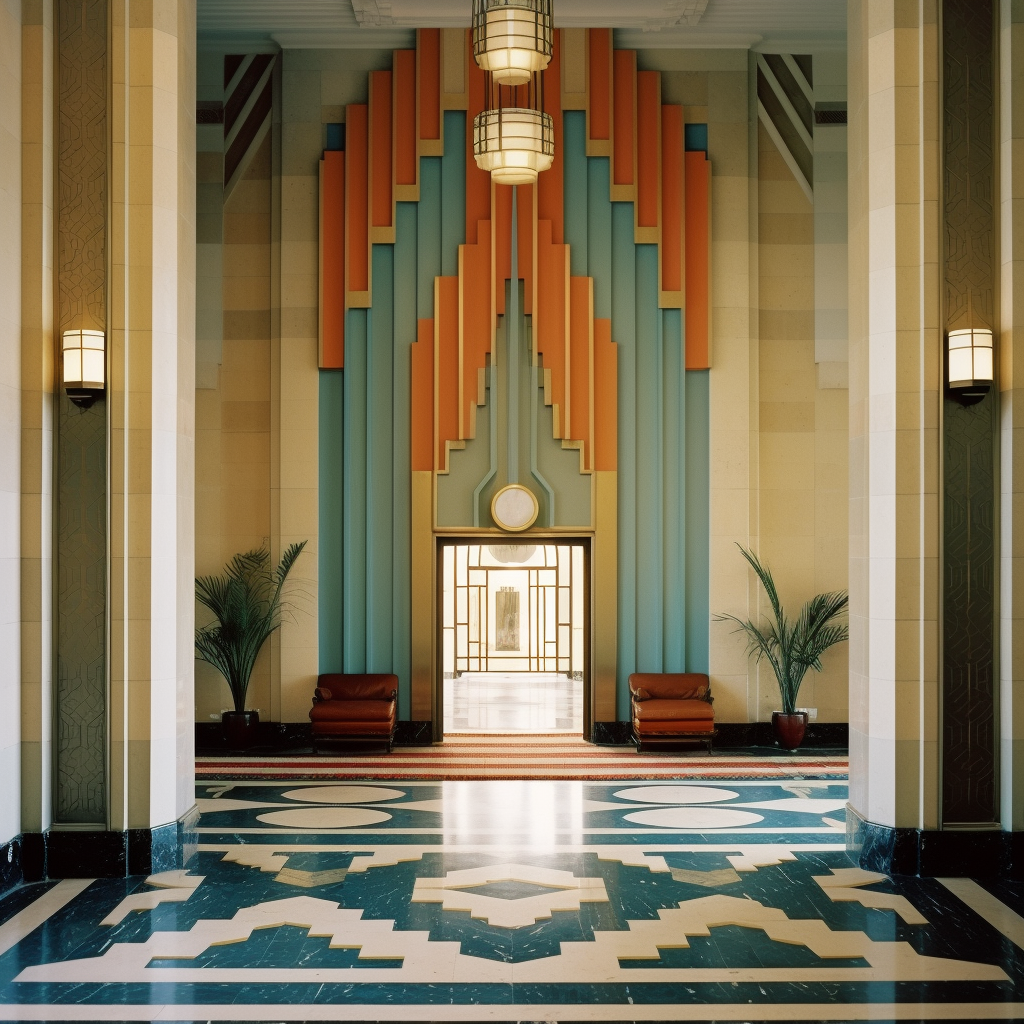}
    \caption{An image generated based on the Art Deco style.}%
    \label{fig:example_art_deco}%
\end{figure}

One of the images generated for this style is shown in Fig.~\ref{fig:example_art_deco}. Wall and floor art are important aspects of this style, thus most of the images do not show an entire building despite being prompted to. They instead show interiors, entrances or windows, most likely because this is what most images about art deco focus on.

Midjourney then generated the following descriptions when this image was fed back into the model. Highlights in \colorbox{blue!30}{Blue} mark direct occurrences of the style name, highlights in \colorbox{green!30}{green} mark occurrences of the description terms (or very similar words), and \colorbox{red!30}{red} mark erroneous style names or terms (unused in this example).

\begin{enumerate}
    \item \enquote{an elaborate mosaic tile lobby tile, in the style of light orange and teal, \colorbox{blue!30}{art deco} \colorbox{green!30}{geometric} \colorbox{green!30}{patterns}, analog photography, spectacular backdrops, confessional, serene atmosphere}
    \item \enquote{entrance to durban conference center in sydney, in the style of \colorbox{blue!30}{art deco} \colorbox{green!30}{geometric} \colorbox{green!30}{patterns}, dark turquoise and light orange, interior scenes, kodak portra, romantic interiors, mediterranean-inspired}
    \item \enquote{a bright hallway with \colorbox{green!30}{colored} art murals, in the style of \colorbox{blue!30}{art deco} \colorbox{green!30}{geometric} \colorbox{green!30}{shapes}, kodak portra, light orange and dark aquamarine, monumental architecture, \colorbox{green!30}{symmetrical} design, grandiose architecture, \colorbox{blue!30}{art deco}}
    \item \enquote{a big entry door, in the style of \colorbox{blue!30}{art deco} \colorbox{green!30}{geometric} \colorbox{green!30}{shapes}, light orange and dark cyan, \colorbox{green!30}{symmetrical} arrangements, kodak ektachrome, expansive spaces, marble, native australian motifs}
\end{enumerate}

We can see that despite the terms in the original prompts, the Midjourney descriptions do not use most of them, and instead reuse many terms that the model apparently learned to associate with the style internally. It has a very clear idea of the styles, made apparent by the fact that every description directly references the style, one of them even does so twice. The data bears this out too, as Fig.~\ref{fig:midjourney_describe_misattr} shows that there are barely any misattributions for Art Deco---the rare exception being Art Nouveau.

\begin{figure}[h]
    \centering
    \includegraphics[width=0.5\textwidth]{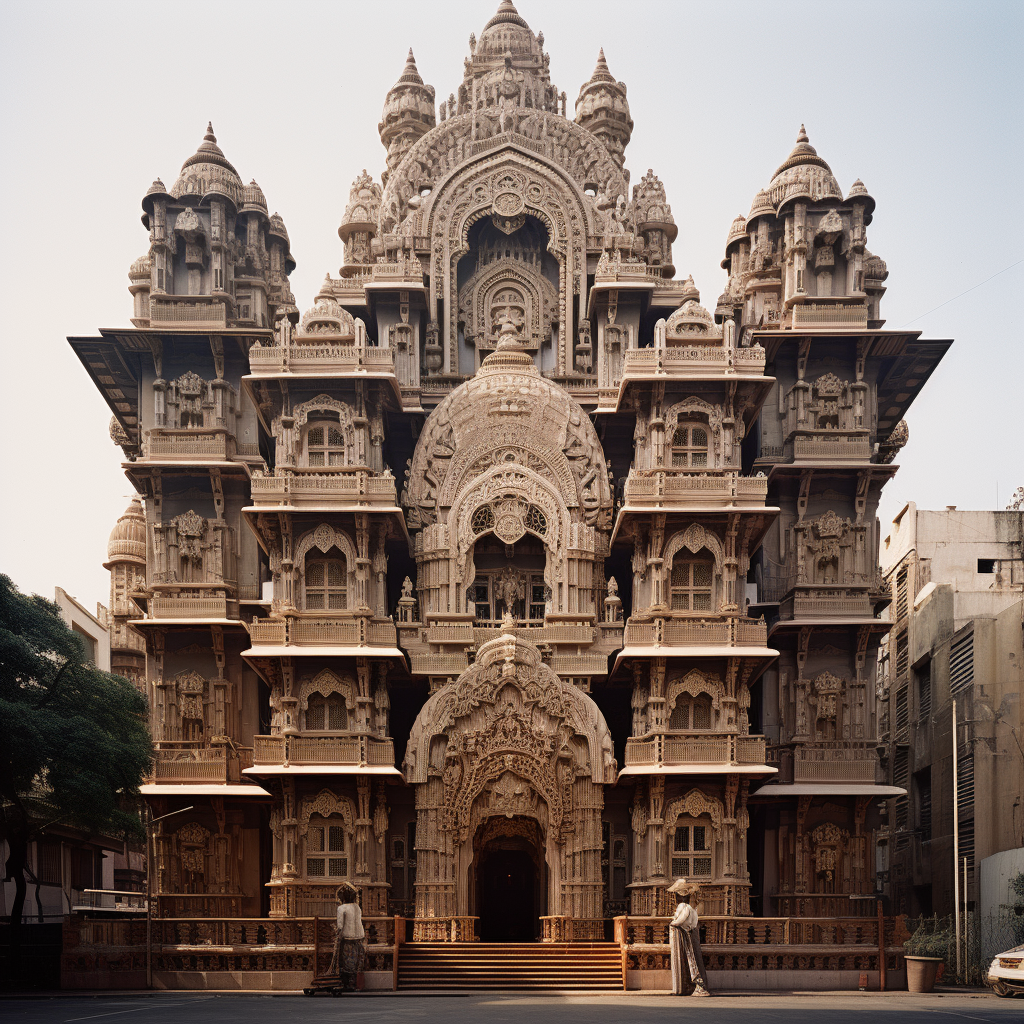}
    \caption{An image generated based on the Gupta style.}%
    \label{fig:example_gupta}%
\end{figure}

We see a different picture with the Gupta style. Gupta architecture is an Indian architecture style mostly known for its filigreed temples. ChatGPT described it with the following terms: \enquote{ornate intricate majestic symmetrical elaborate exquisite grandiose opulent regal lavish Sanskritized Indian Hindu Buddhist Jain ancient magnificent}. An image that Midjourney generated for this prompt is shown in Fig.~\ref{fig:example_gupta}.

Midjourney then in turn tried to describe this image with the following four prompts:

\begin{enumerate}
    \item \enquote{an old and very \colorbox{green!30}{elaborate} \colorbox{green!30}{indian} temple outside of city, in the style of dark beige and gray, contax/yashica mount, futuristic architecture, \colorbox{green!30}{intricately} sculpted, design by architects, flickr, commission for}
    \item \enquote{a typical \colorbox{green!30}{indian} temple structure sitting amongst a city street, in the style of, \colorbox{green!30}{symmetrical} forms, dark beige and gray, restored and repurposed, carved religious icons, contax tix}
    \item \enquote{\colorbox{green!30}{india's} largest mausoleum a temple built by gujarat state government, in the style of decorative \colorbox{red!30}{art nouveau}, dark beige and gray, sculptural grotesqueries, adox silvermax, 1970-present, confessional, richly layered}
    \item \enquote{an old temple, with statues and statuettes of saints, in the style of \colorbox{green!30}{intricate} cut-outs, dark beige and silver, monumental architecture, yashica t4, \colorbox{green!30}{symmetrical} forms, \colorbox{green!30}{intricate} \colorbox{red!30}{art nouveau}}
\end{enumerate}

It is immediately apparent that there is no mention of the word Gupta, and instead there are two misattributions to the Art Nouveau style, which seem far fetched based on the image. The reason this misattribution could be taking place is the intricate level of decorative detail that both styles often feature. Other identifying words like \enquote{contax tix, yashica t4, adox silvermax} are related to the photography itself and not the style or specific building design. 

These two styles exemplify the results from the previous section quite well: some styles can rely on being clearly encoded through their name---for others, longer and more specific descriptions are needed in order for Midjourney to generate appropriate images.

\subsubsection{Encoded Motifs}

\begin{figure}[t]%
    \centering
    \subfloat[\centering Gothic with description terms]{{\includegraphics[width=0.40\textwidth]{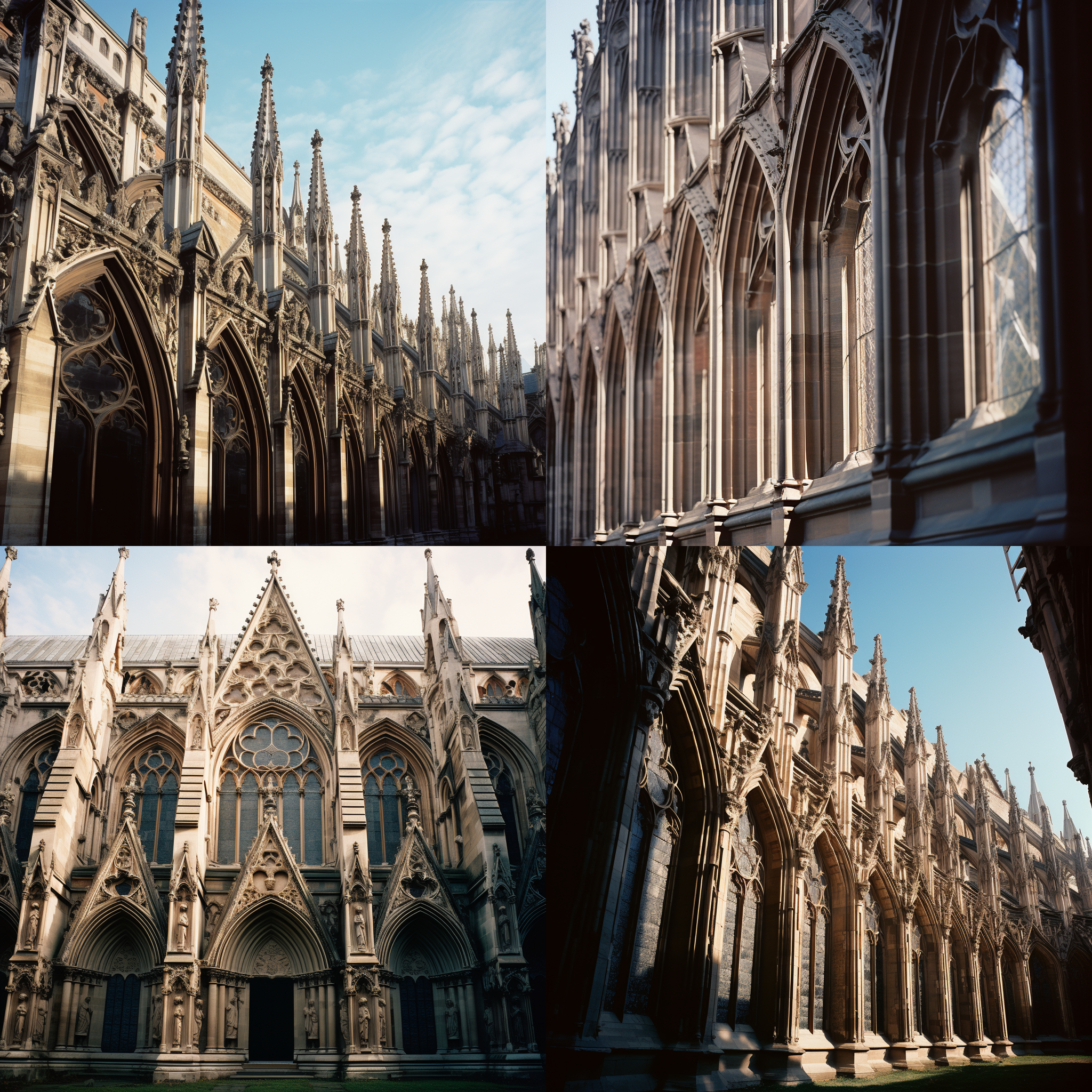} }}%
    \subfloat[\centering Gothic without description terms]{{\includegraphics[width=0.40\textwidth]{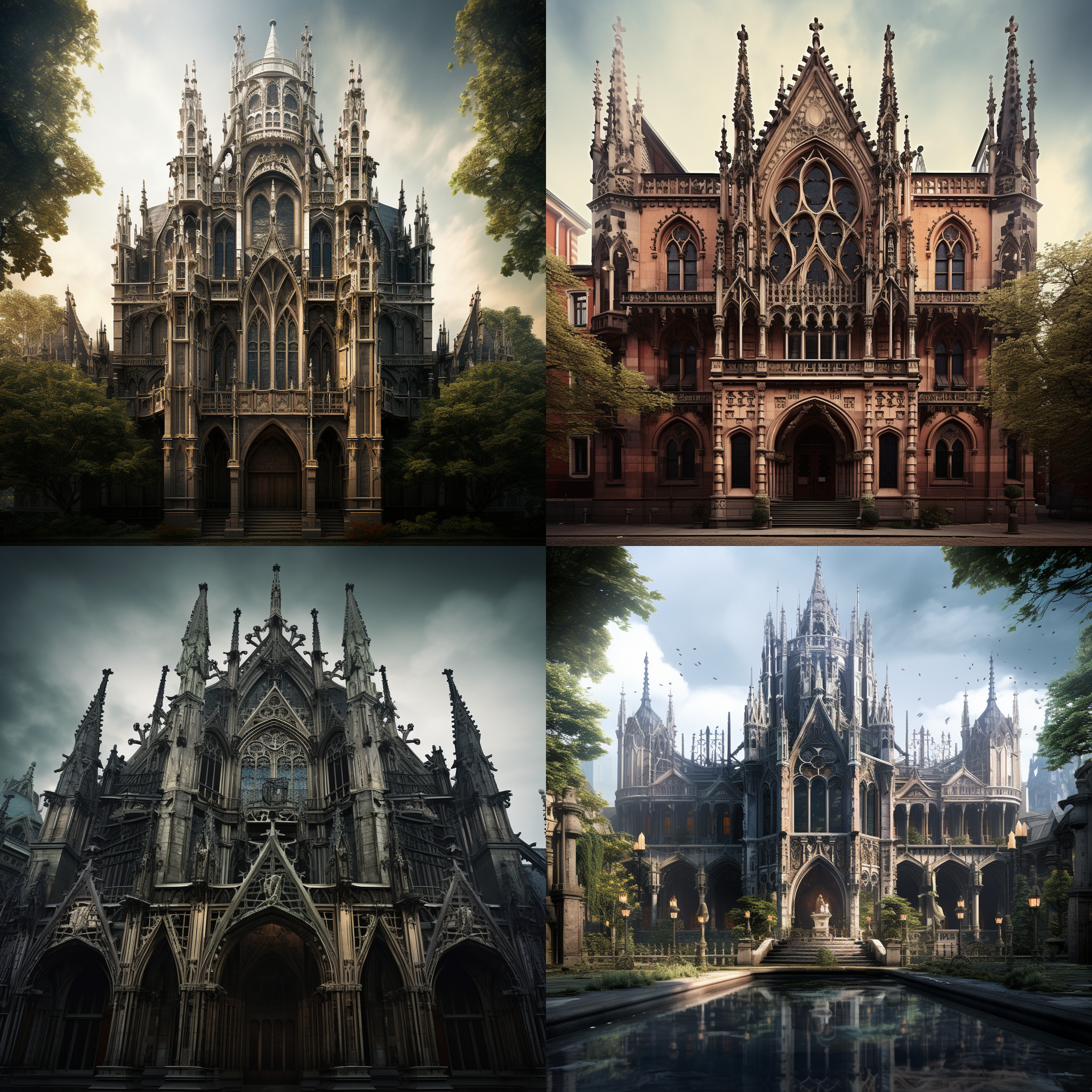} }}%

    \subfloat[\centering Ancient roman with description terms]{{\includegraphics[width=0.40\textwidth]{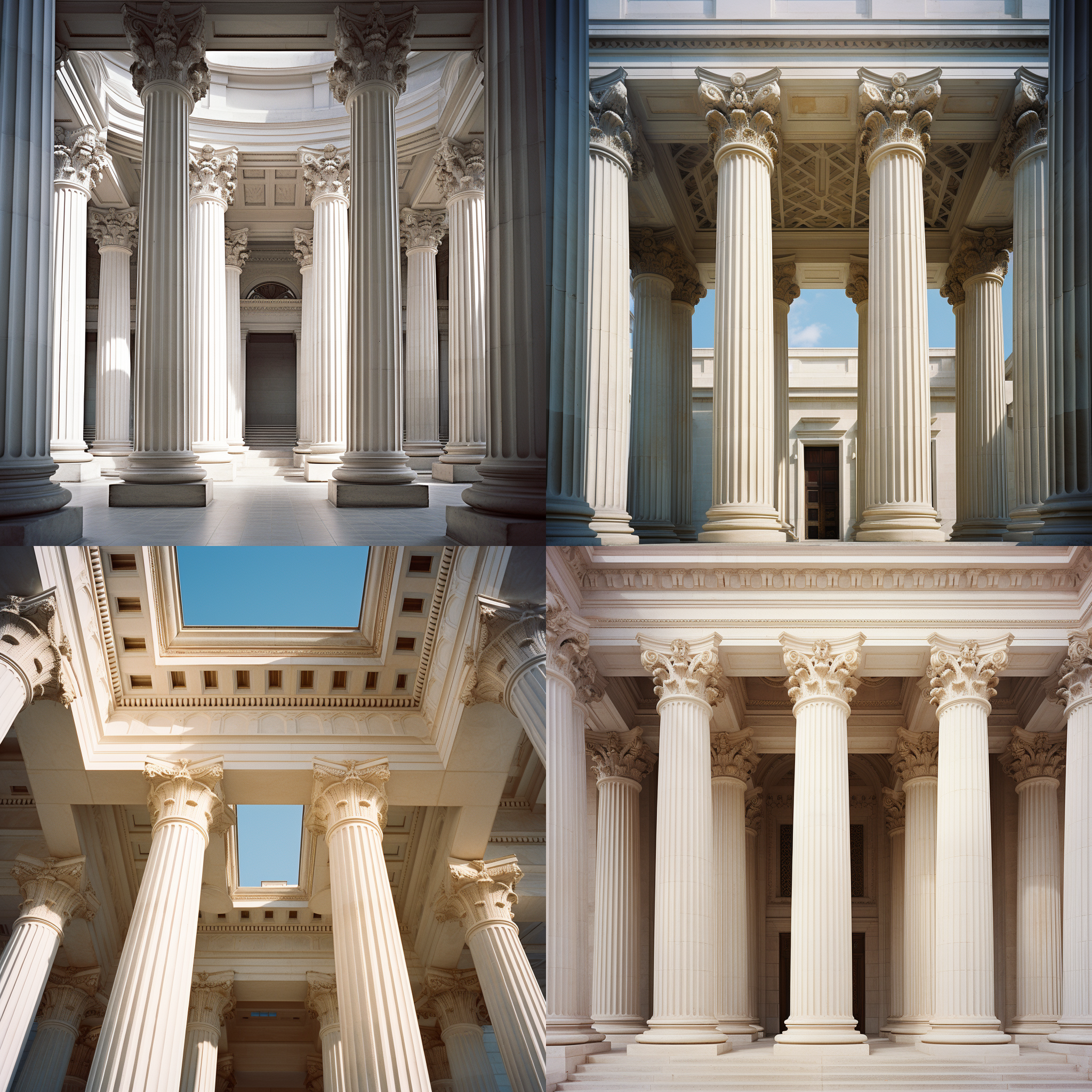} }}%
    \subfloat[\centering Ancient roman without description terms]{{\includegraphics[width=0.40\textwidth]{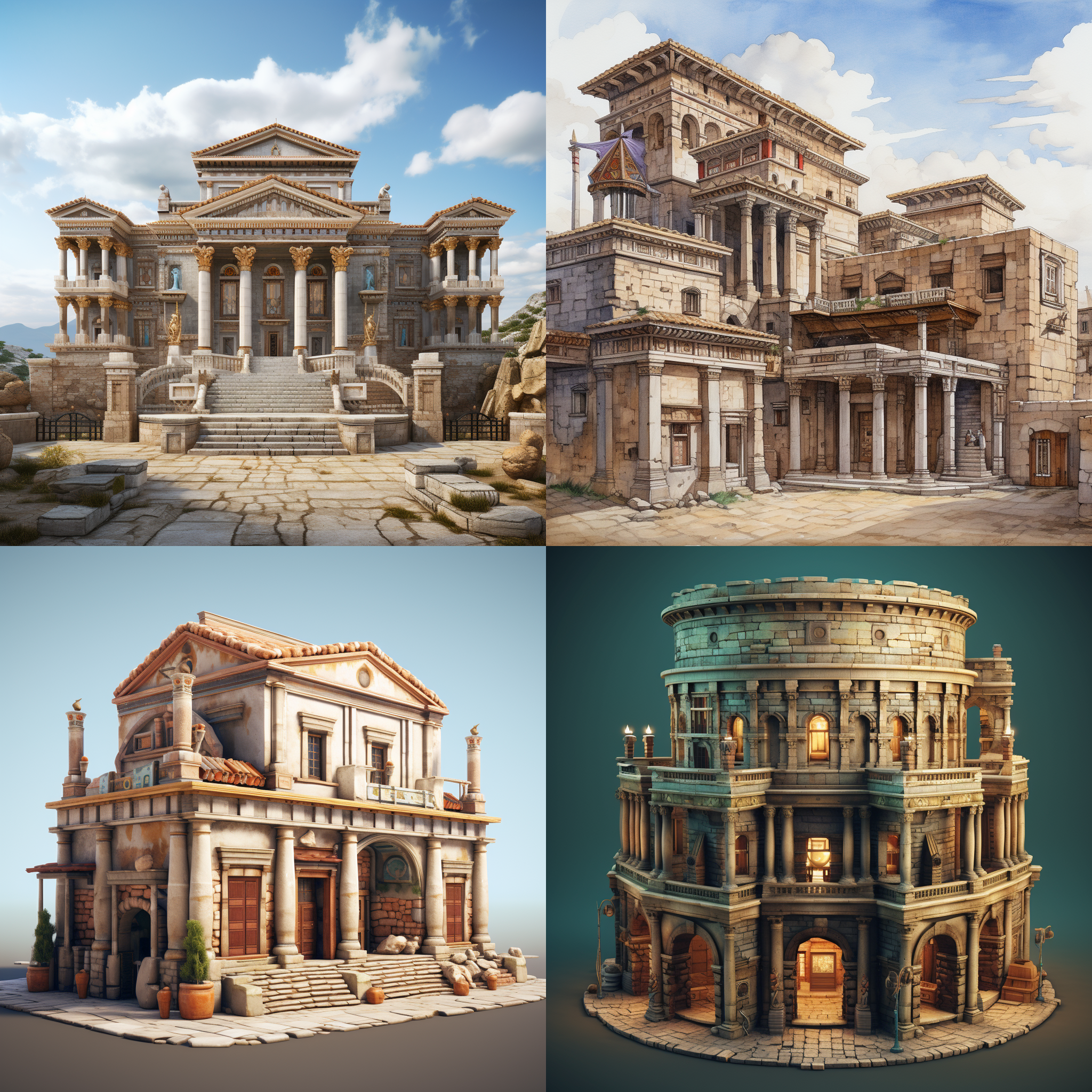} }}%
    \caption{Differences in how iconic motifs are baked into the styles either within Midjourney and within ChatGPTs description terms, for the example of two styles.}%
    \label{fig:motif_encoding}%
\end{figure}

While Midjourney succeeds in rendering all styles in vivid and photographic detail, differences in quality are more subtle. For example, for some styles, certain iconic motifs appear in the majority of the image outputs, unless explicitly prompted otherwise. Some of this could stem from ChatGPT, specifically from the list of style descriptors that we included in the prompts, but this also occurs when only using Midjourney.

To demonstrate this, in Fig.~\ref{fig:motif_encoding} we can see how two styles behave with and without the additional terms describing common style elements. In both cases, we see that the terms create a selection of very similar results, because desired elements are named in the prompt. How well these images represent their individual style is somewhat subjective, but it is also a measure of how broad and well ChatGPT was able to describe each individual style.

Perhaps more interesting is the distinction between \ref{fig:motif_encoding}b and \ref{fig:motif_encoding}d. Without terms, the gothic style still follows a very distinct motif, invoking images that are culturally the most present for this style. The roman style is far more varied, producing results that are more dissimilar in both motif and rendering style. In combination with the earlier quantitative results, this seems to be an effect of how clear the picture of a style is within Midjourney's model. For a style that the model can already describe well, the terms do not change much, while the less well described styles are more nebulous and create a larger variety in the images without the additional description terms.

For certain applications, the greater variety could be an advantage. For historical applications, we would usually want the model to have a very clear and realistic idea of the style. Broadness might be a marker of quality in the case of ChatGPT, where we want an encompassing and not too specific description of each style, while with Midjourney, we require a very specific visual identity for each style.

\subsubsection{Encoded Biases}

\begin{figure}[h]
    \centering
    \includegraphics[width=0.9\textwidth]{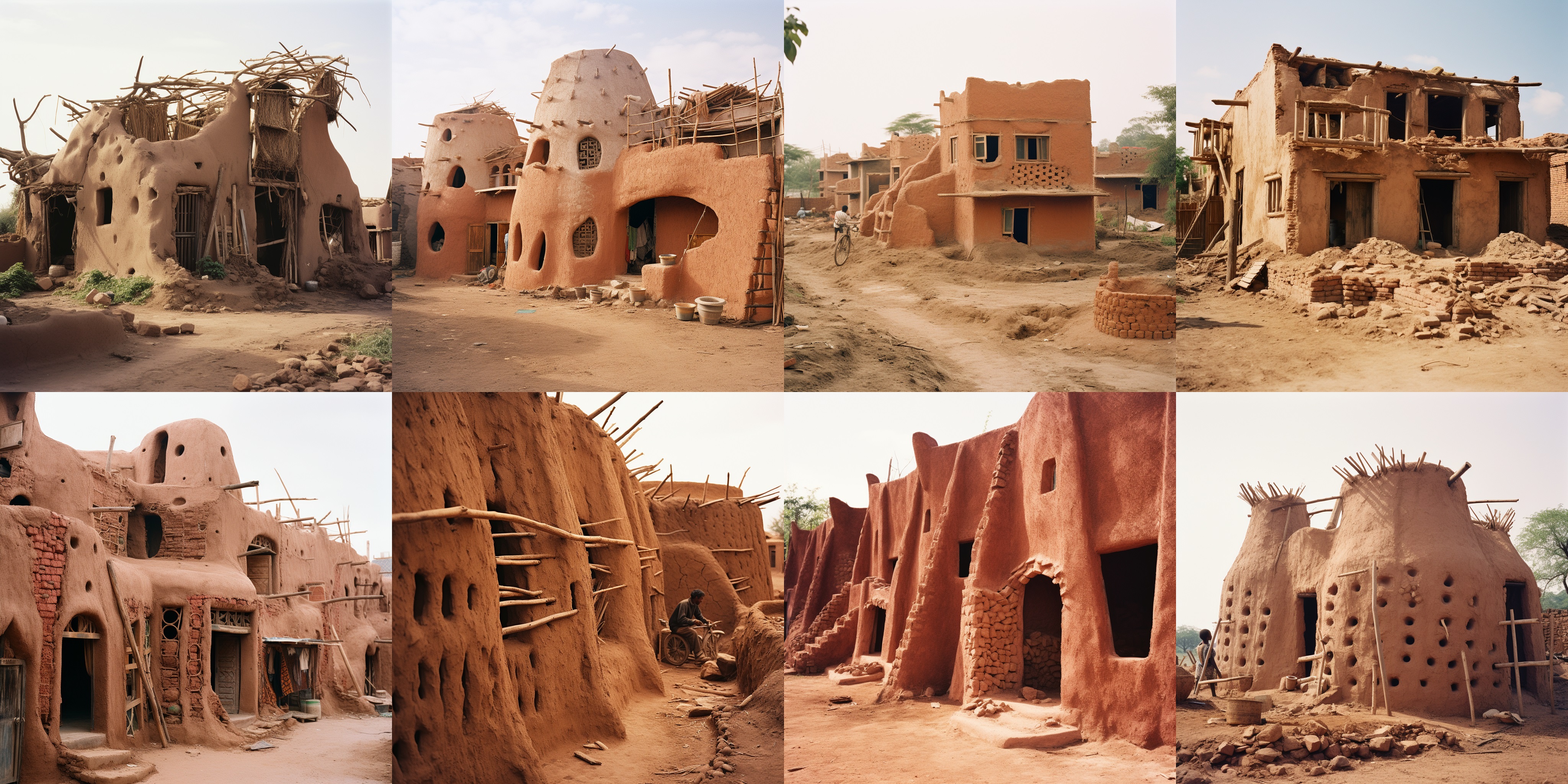}
    \caption{Eight image variants for the Mud Brick architecture query.}%
    \label{fig:bias_mudbrick}%
\end{figure}

\begin{figure}[h]
    \centering
    \includegraphics[width=0.9\textwidth]{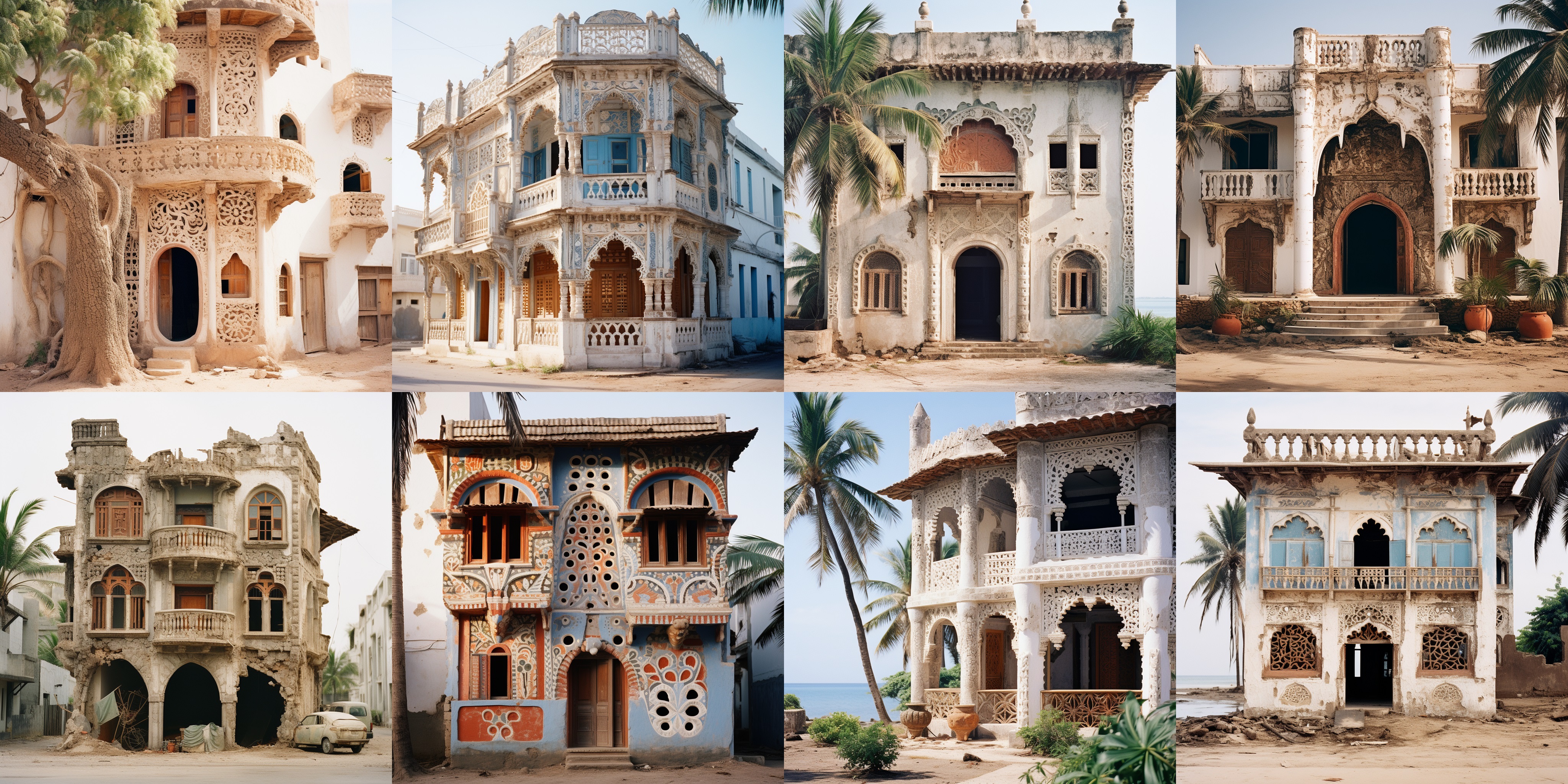}
    \caption{Eight image variants for the Swahili architecture query.}%
    \label{fig:bias_swahili}%
\end{figure}

While much work has been done to reduce harmful biases from the models throughout the several past version updates, our analysis still shows clear issues. Common narratives from movies and news influence the generation results in very immediate, as well as more subtle ways. Two striking instances of this shall be named here. None of the following aspects were asked for in the prompts:

\begin{enumerate}
    \item Mud Brick architecture prompts create humans far more frequently than other styles, instead of only showing the building and its architecture. There are also frequent appearances of building ruins or unfinished structures, while clearly not depicting historic artifacts, as shown in Fig.~\ref{fig:bias_mudbrick}.
    \item Swahili architecture often includes cars and bikes within the generations. More problematically however, it is also far more likely to show damaged or desolate structures, with some of the imagery invoking elements of conflict damage. Fig.~\ref{fig:bias_swahili} shows the result of two Midjourney requests. This is despite ChatGPT's description including positive terms like \enquote{vibrant}.
\end{enumerate}

This bias is likely an effect of international, and especially western, news narratives heavily influencing the content of the model, as many of the training images for such styles can be assumed to be taken from news stories and documentaries, instead of from historical records or architectural databases.

\section{Analysis of Architectural Queries in Use}\label{sec:inuse}

\begin{figure}[t]%
    \centering
    \subfloat[\centering Word Frequency\label{fig:word_freq_word}]{{\includegraphics[width=0.49\textwidth]{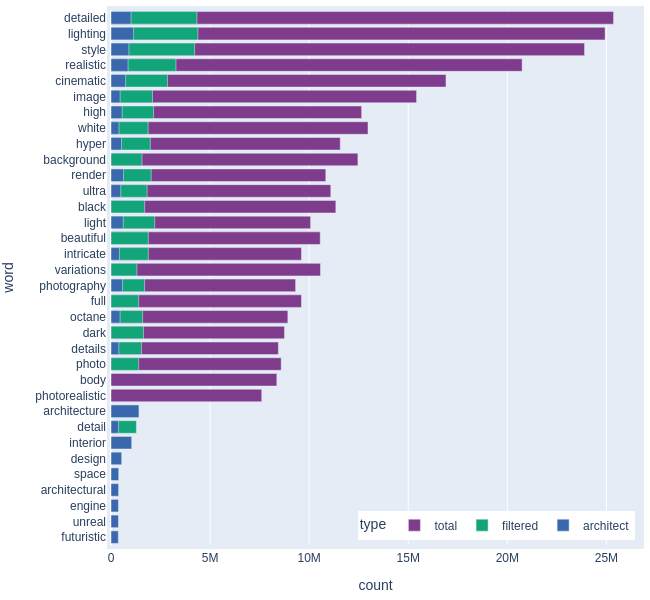} }}%
    \subfloat[\centering Keyword Frequency\label{fig:word_freq_kword}]{{\includegraphics[width=0.49\textwidth]{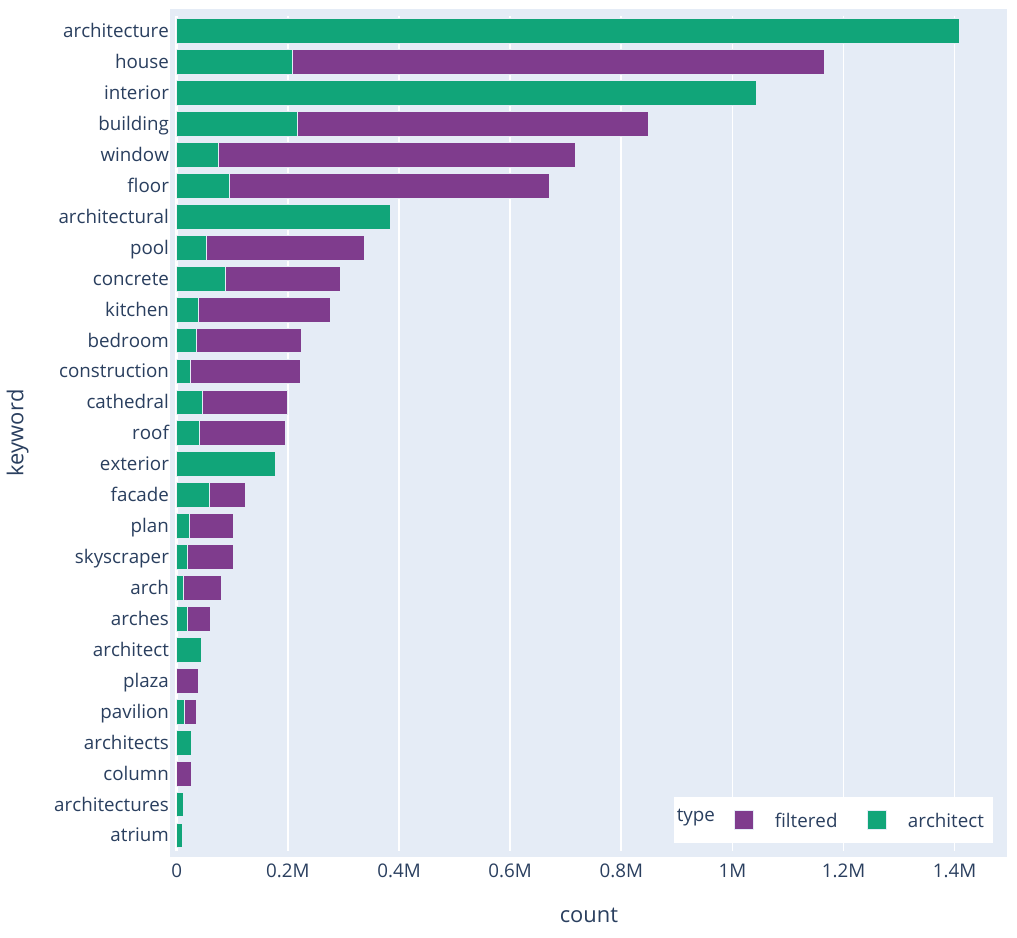} }}%
    \caption{Word and Keyword Frequency in Midjourney}%
    \label{fig:word_freq}%
\end{figure}

We also explored how people use these AI art platforms in practice. We base our analysis on about \totalsize queries that users posted on Midjourney. Those prompts are publicly visible, as Midjourney is using the Discord messaging app as its main interface, which allowed us to monitor the public channels for queries that we consider to be of an architectural nature~\cite{ploennigs2023ai}. 
We selected queries containing either the word \enquote{architect}, \enquote{interior} or \enquote{exterior} design or one of 38 architectural keywords like \enquote{building}, \enquote{facade}, or \enquote{construction}  (listed in Fig.~\ref{fig:word_freq} (b) and \footnote{Keywords not listed in Fig.~\ref{fig:word_freq} (b) due to low count: balcon, basilica, battlement, buttress, gable,  hvac, latticework, livingroom, minaret, panelling, pavilion, plinth, rotunda, spire}). We identified these keywords by selecting only those from architectural glossaries\footnote{\url{https://www.heritage.nf.ca/articles/society/architectural-terms.php}}$^,$\footnote{\url{https://en.wikipedia.org/wiki/Glossary_of_architecture}} that co-occurred in at least 10\,\% of all cases with \enquote{architect}, \enquote{interior} or \enquote{exterior} in the queries. We also added to the list of keywords the names of 941 famous architects from Wikipedia\footnote{\url{https://en.wikipedia.org/wiki/List_of_architects}} as we noted that several queries actually refer to the style of these architects. By applying these filters, we identified \filtersize queries (11.54\,\%) with potential architectural intent including \archsize queries (2.6\,\%) explicitly containing \enquote{architect}, \enquote{interior} or \enquote{exterior} design.

In the next steps we filter out stopwords and build a Word2Vec model  \cite{mikolov2013distributed} from these queries to get a model of the occurrence and co-occurrence of terms in these queries. 
For understanding the results, it is important to know that most Midjourney users do not formulate full sentences, but, rather a collection of terms that refer to the content, style, or render quality of the targeted image.
Fig.~\ref{fig:word_freq_word} shows the  most frequent terms used in the filtered queries. The color purple represents the frequency across all \totalsize queries, green is the frequency within the filtered \filtersize queries and blue within \archsize queries explicitly containing \enquote{architect}, \enquote{interior} or \enquote{exterior} design. The top 10 has a similar frequency across all three classes. Many of those refer to Midjourney style commands like \enquote{detailed}, \enquote{realistic}, \enquote{cinematic}, \enquote{render}. However some terms like \enquote{black}, \enquote{full} or \enquote{portrait} have high overall frequency, but, low frequency in architectural context. Other terms like \enquote{architecture}, \enquote{interior}, \enquote{house}, \enquote{building} do only occur exclusively within our filtered results as they are part of our keyword list.

Fig.~\ref{fig:word_freq_kword} lists these keywords that we use and their respective frequency. As we filter on these keywords, their total frequency is identical with the filtered one and we do not display it. It is of note that \enquote{architecture} and \enquote{interior} keywords are the most and third frequent keyword. It is notable that \enquote{interior} is six times more popular than \enquote{exterior} design as keyword, but this simply may be that users refer to it implicitly through \enquote{architecture}.

\begin{figure}[t]%
    \centering
    
    \subfloat[\centering Query Length. x-axis labels show the \# of queries.\label{fig:query_length_q}]{{\includegraphics[width=0.49\textwidth]{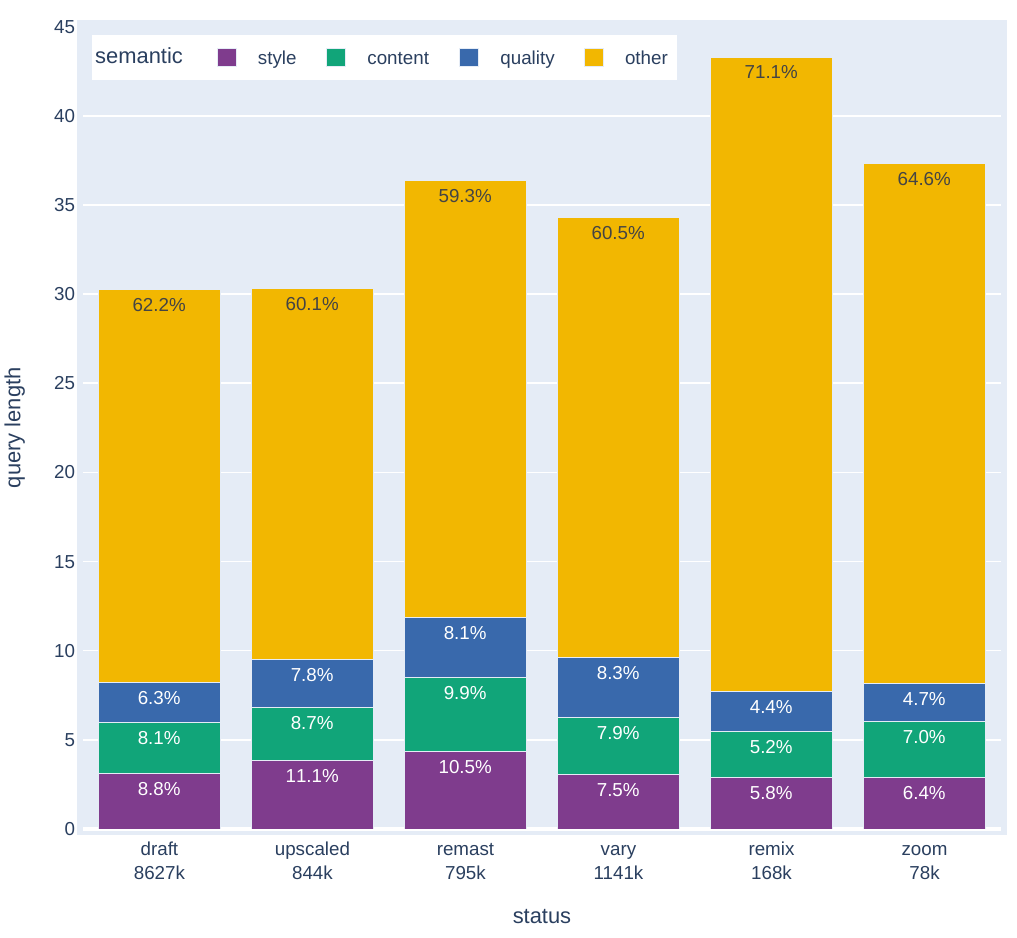} }}%
    \subfloat[\centering Query Workflows. x-axis labels show the \# of flows.\label{fig:query_length_wf}]{{\includegraphics[width=0.49\textwidth]{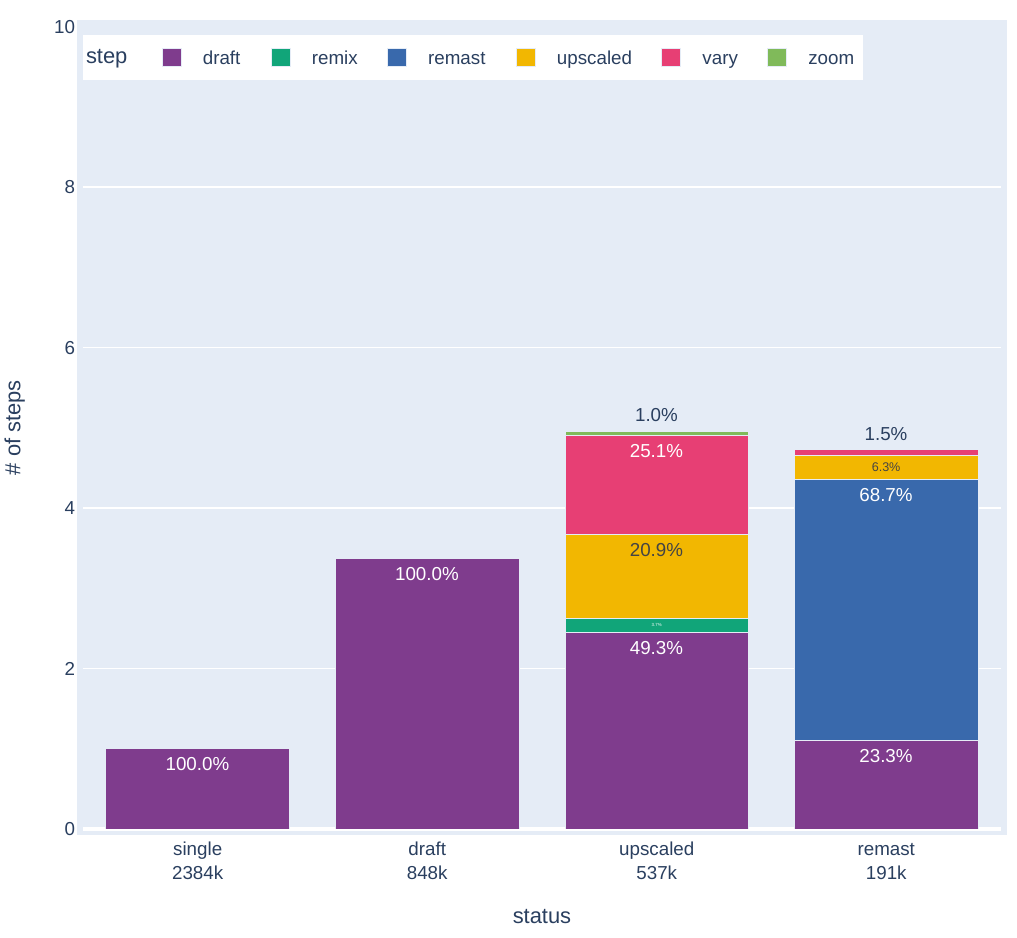} }}%
    \caption{Query and workflow length in Midjourney}%
    \label{fig:query_length}%
\end{figure}

Fig.~\ref{fig:query_length_q} shows the mean length of queries depending on whether they got upscaled, remastered or left in draft mode. A draft mode image is of low image size, so users will normally upscale or remaster them if they like one of the variants. It is notable that for the remastered options and the remixed version the mean query length increases to 37 and 43 terms per query in comparison to 30 terms for draft mode queries. We also manually classified the most frequent 150 terms into the categories: style, content, quality. It is notable that for the remastered queries, the percentage of style terms increases significantly.

Fig.~\ref{fig:query_length_wf} shows the mean number of iterations needed to develop a query. We classify a query as iteration if the same user is rerunning the same or extended query within 30 minutes. 54\,\% of all unique queries (single, 852k) are run only once. The other half of the queries are improved in multiple iterations. Queries that remain in draft mode require about 3.7 steps. 7.8\,\% of the queries are good enough to be upscaled require about 6.75 steps in total. They are upscaled after 4.1 draft steps into different variants (light, medium/beta, max). 5.3\,\% queries that are remastered take about 5.1 iterations. They have only 1.4 draft mode queries, but 1.2 remastering steps, and 2.3 final upscale steps.

This illustrates that users do not come-up with perfect queries from scratch, but normally develop them over multiple iterations by selecting the best variants or adding more terms (especially style terms). From these insights as well as practical experience we developed some recommended workflows that we will discuss in the next section.

\begin{figure}[t]%
    \centering
    \subfloat[\centering Famous Architect Frequency\label{fig:archi_freq_arch}]{{\includegraphics[width=0.49\textwidth]{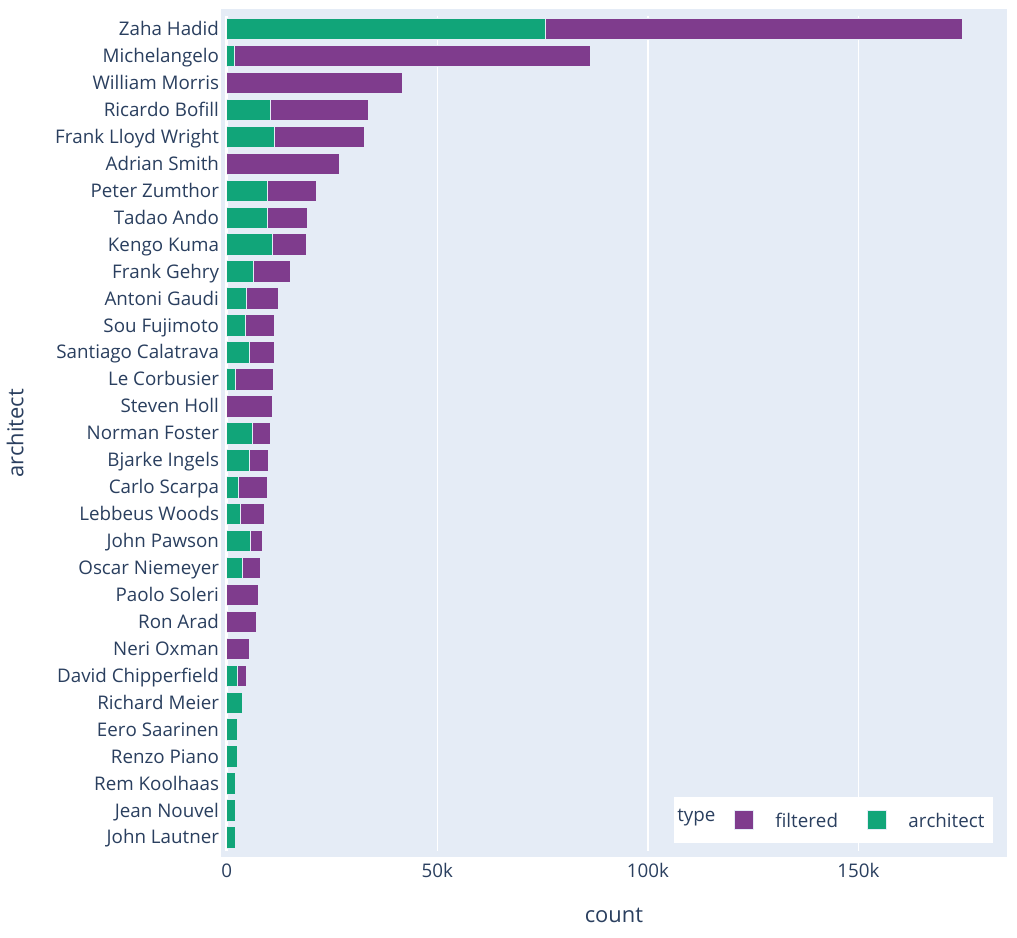} }}%
    \subfloat[\centering Style Frequency\label{fig:archi_freq_style}]{{\includegraphics[width=0.49\textwidth]{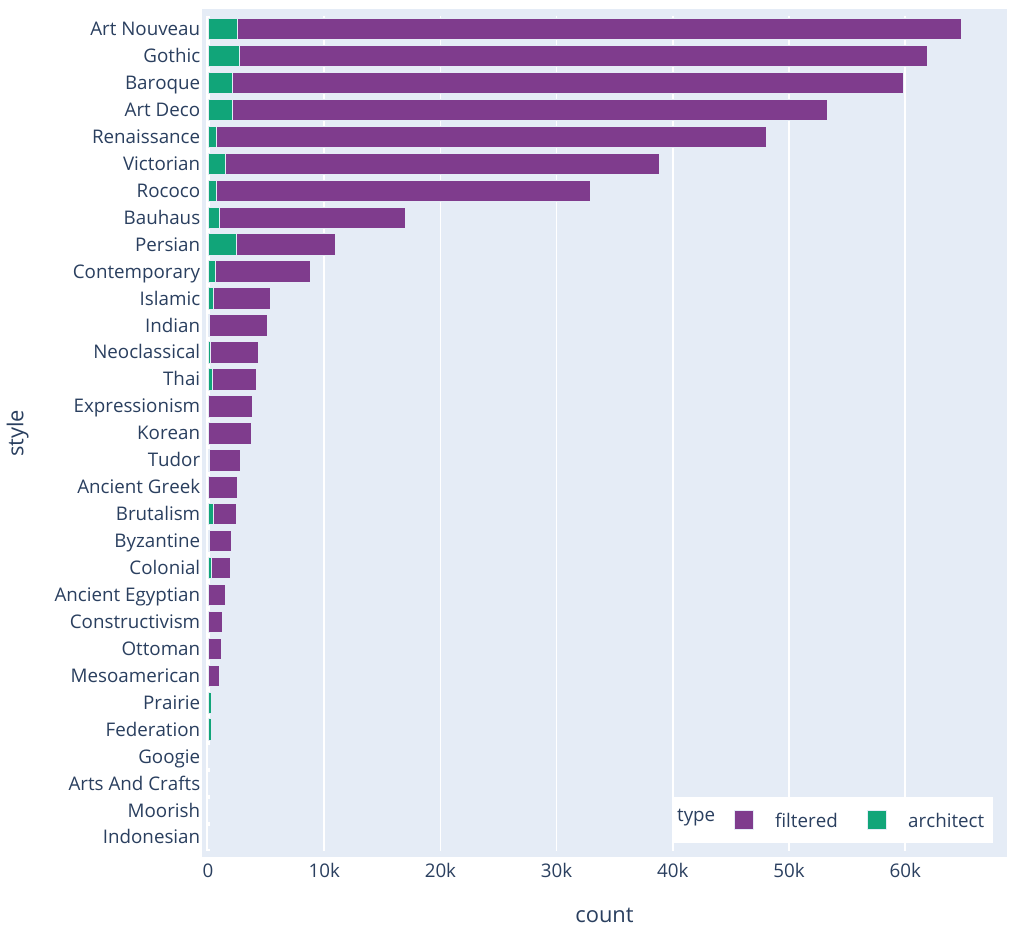} }}%
    \caption{Architect and Style Frequency in Midjourney}%
    \label{fig:archi_freq}%
\end{figure}

Fig.~\ref{fig:archi_freq_arch} shows the frequency of famous architects that we extracted from queries that referred to at least one of them. Zaha Hadid is by far the most frequently queried architect, given her well recognizable style and the likely popularity of her designs within the same social groups that are interested in experimenting with AI tools. Michelangelo is second, usually invoked for his art style instead of his architectural contributions, similar to William Morris. This is indicated by the very low green bar, meaning that the name may be frequent in all results related to architecture, but, rarely directly referencing architecture, interior or exterior design. The architects Ricardo Bofill, Frank Lloyd Wright, Adam Smith, Peter Zumthor, Tadao Ando, Kango Kuma, Frank Gehry, and Antoni Gaudi complete the top 10. They are often used in an explicitly architectural context beside Adam Smith.

Fig.~\ref{fig:archi_freq_style} shows which architectural styles are used most frequently in the queries that referred to at least one style. \enquote{Art Nouveau} is the most frequently used style followed by \enquote{Gothic}, \enquote{Baroque}, and \enquote{Art Deco}. This is partially driven by the popularity of those styles in pop art, as the green bar is quite low in comparison to Fig.~\ref{fig:archi_freq_arch}. Interestingly, these are also styles for which Midjourney shows a good capability of recognizing them by name (see Fig.~\ref{fig:midjourney_describe_occur} and discussion). In the middle we have \enquote{Renaissance}, \enquote{Victorian}, and \enquote{Rococo}. Among all those styles we do not find a single modern style. Bauhaus is on position 8 and Contemporary on position 10, with significantly lower frequency of use. Fig.~\ref{fig:archi_freq_style} also lists styles out of the full list of 111 originally extracted styles from ChatGPT that we did not investigate in our previous analysis, to show that the original prioritization retrieved with ChatGPT is not bad, at least in terms of usage. 

\begin{figure}[t]%
    \centering
    \subfloat[\centering Architect by Style\label{fig:archi_styles_freq}]{{\includegraphics[width=0.49\textwidth]{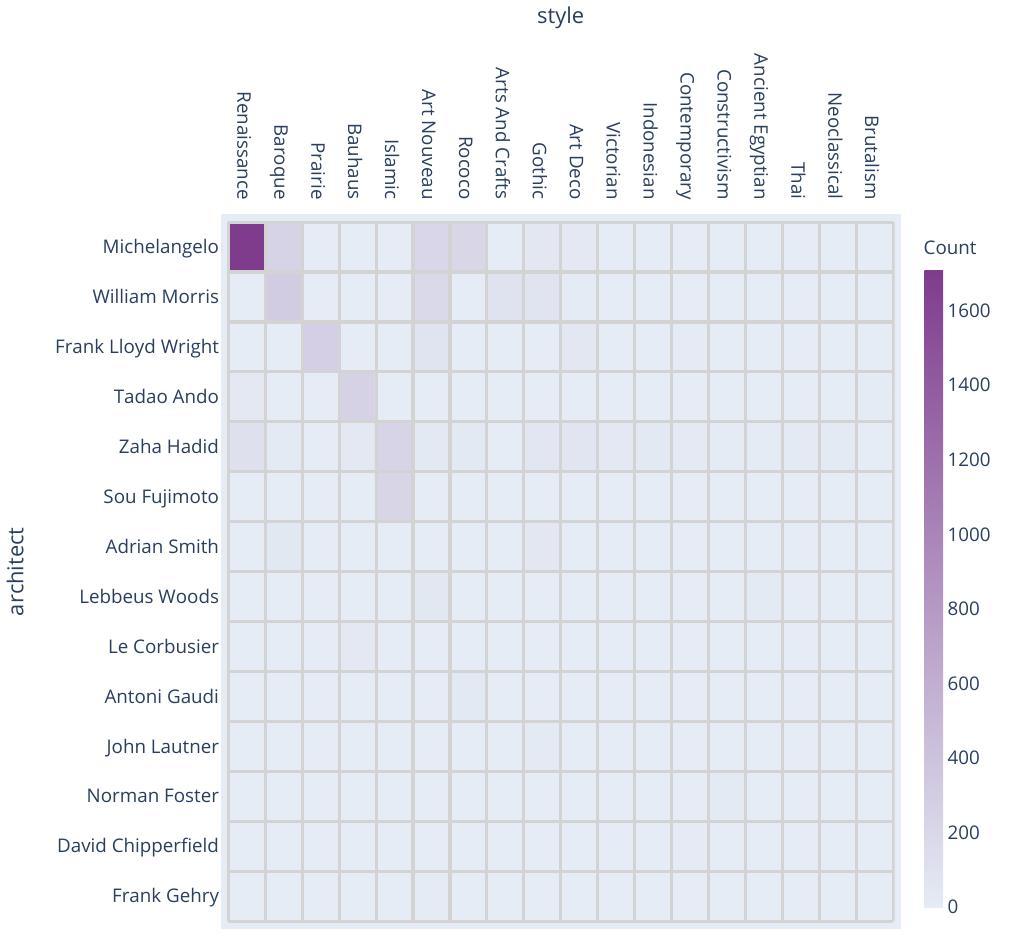} }}%
    \subfloat[\centering Architect Styles combinations\label{fig:styles_styles_freq}]{{\includegraphics[width=0.49\textwidth]{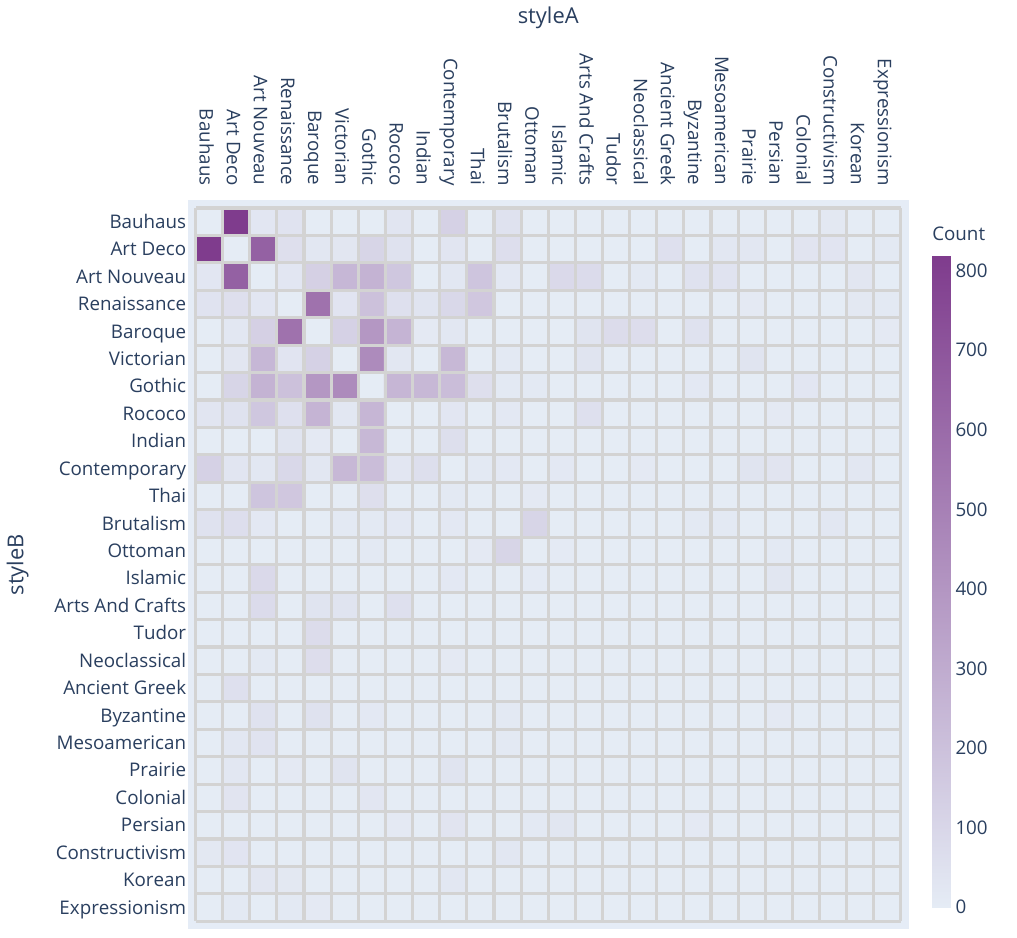} }}%
    \caption{Architect by Style Frequency}%
    \label{fig:archi_styles_freq}%
\end{figure}

Fig.~\ref{fig:archi_styles_freq} shows how frequent the architects are queried together with a specific architectural style. Michelangelo is used with the \enquote{Renissance} style in 3.6\,\% of all usages of the style. William Morris with \enquote{Baroque}. Zaha Hadid and Sou Fujimoto are most frequently used with \enquote{Islamic} style.

Fig.~\ref{fig:styles_styles_freq} shows which styles are used in combination. 
The most frequently combination is \enquote{Bauhaus} with \enquote{Art Deco} followed by \enquote{Art Nouveau} with \enquote{Art Deco}. Here, it is to note that style combinations occur less frequent as the maximum number is 818 combinations, which is 1.5\,\% of the occurrences of \enquote{Art Deco}. Latter correlates with the high frequency of use of both styles as shown in Fig.~\ref{fig:archi_freq_style}. Interestingly, \enquote{Bauhaus} is less frequently used, but, the combination with \enquote{Art Deco} appears to be a favorite. The style with the most diverse styles combinations is \enquote{Gothic}. It is combined with \enquote{Victorian} and \enquote{Baroque} architecture, but, also with \enquote{Contemporary}, \enquote{Indian}, \enquote{Art Nouveau} and \enquote{Renaissance}. Which also occurs with \enquote{Baroque}. The other styles are even less often combined.



\section{Conclusion}\label{sec:conclusion}

Throughout this chapter we explored how historical styles are represented in generative AI tools and how they are used in practice. This gives us new insights into both the kinds of visual characteristics current AI models tend to perceive and reconstruct with architectural styles, and into how these styles are embedded in our own cultural consciousness by how a global user base employs them for art generation. Our analyses showed that while all styles are in some way present in the models and can be queried, there are large differences in quality, especially along cultural and temporal boundaries. 

Our investigation into the architectural knowledge in ChatGPT 3.5 revealed that it actually has good knowledge of a large cultural variety of styles. It can explain and differentiate them well, as long as general descriptions and characteristics are requested. It is less reliable in listing relevant architects and examples, with a tendency to hallucinate new architects based on news mentions, like for presidents that opened a building. But, this can usually quickly be alleviated by simply asking control questions back to ChatGPT, or by averaging across generations variants.

We could also confirm that Midjourney has a robust knowledge of many architectural styles, with a clear bias towards the kinds of styles that are of specific interest to its users. When targeting lesser-known or non-western styles, we see that the highest level of quality is achieved by combining ChatGPT and Midjourney in order to generate description terms first, and then using them in the text-to-image prompts. This is partially related to Midjourney's behaviour that more specific and longer prompts lead to better results, which we confirmed in our analysis of the \totalsize Midjourney queries used in practise.

The analysis of those queries shows us that Midjourney is very frequently used for generating architectural images. In fact, more than 11\,\% of the queries are related to architecture. People often use longer prompts to specify their result image successfully. We also confirmed that in our discussion of Fig.~\ref{fig:motif_encoding} where the prompts with terms usually resulted in more specific images. The architectural styles that people reference are not necessarily the most commonly constructed today, in contrary to the list of the most frequently used architects that is lead by modern architects. 

Some of the issues we identified stem from how the models were trained and could be improved by utilizing more diverse training data sets. In the meantime, users should take extra care in prompting and in using results generated by all types of models: text-to-text, image-to-text and text-to-image. Some problems, like hallucinations, are however inherent to the model architecture. While we likely have to wait for more robust model architectures to fix these problems, our work shows that the current level of accuracy can already be enough for many types of applications.


\bibliographystyle{elsarticle-num} 
\bibliography{bibliography}

\end{document}